\pdfoutput=1

\documentclass[11pt]{article}

\usepackage[]{EMNLP2022}

\usepackage{times}
\usepackage{latexsym}

\usepackage[T1]{fontenc}

\usepackage[utf8]{inputenc}

\usepackage{microtype}

\usepackage{multirow}
\usepackage{graphicx}
\usepackage{enumitem}

\definecolor{green}{rgb}{0.0, 0.5, 0.0}
\usepackage{hyphenat}
\usepackage[perpage]{footmisc}

\newcommand{\stitle}[1]{\vspace{0.5em}\noindent{\bf #1}}
\usepackage{cleveref}
\crefformat{section}{\S#2#1#3}
\crefformat{subsection}{\S#2#1#3}
\crefformat{subsubsection}{\S#2#1#3}
\crefrangeformat{section}{\S\S#3#1#4 to~#5#2#6}
\crefmultiformat{section}{\S\S#2#1#3}{ and~#2#1#3}{, #2#1#3}{ and~#2#1#3}
\usepackage{refstyle}
\Crefformat{figure}{#2Fig.~#1#3}
\Crefmultiformat{figure}{Figs.~#2#1#3}{ and~#2#1#3}{, #2#1#3}{ and~#2#1#3}
\Crefformat{table}{#2Tab.~#1#3}
\Crefmultiformat{table}{Tabs.~#2#1#3}{ and~#2#1#3}{, #2#1#3}{ and~#2#1#3}
\Crefformat{appendix}{#2Appx.~\S#1#3}
\crefformat{algorithm}{Alg.~#2#1#3}
\usepackage{booktabs}
\usepackage{amsfonts}
\usepackage{tcolorbox}
\newcommand*\smiley{\includegraphics[width=1em]{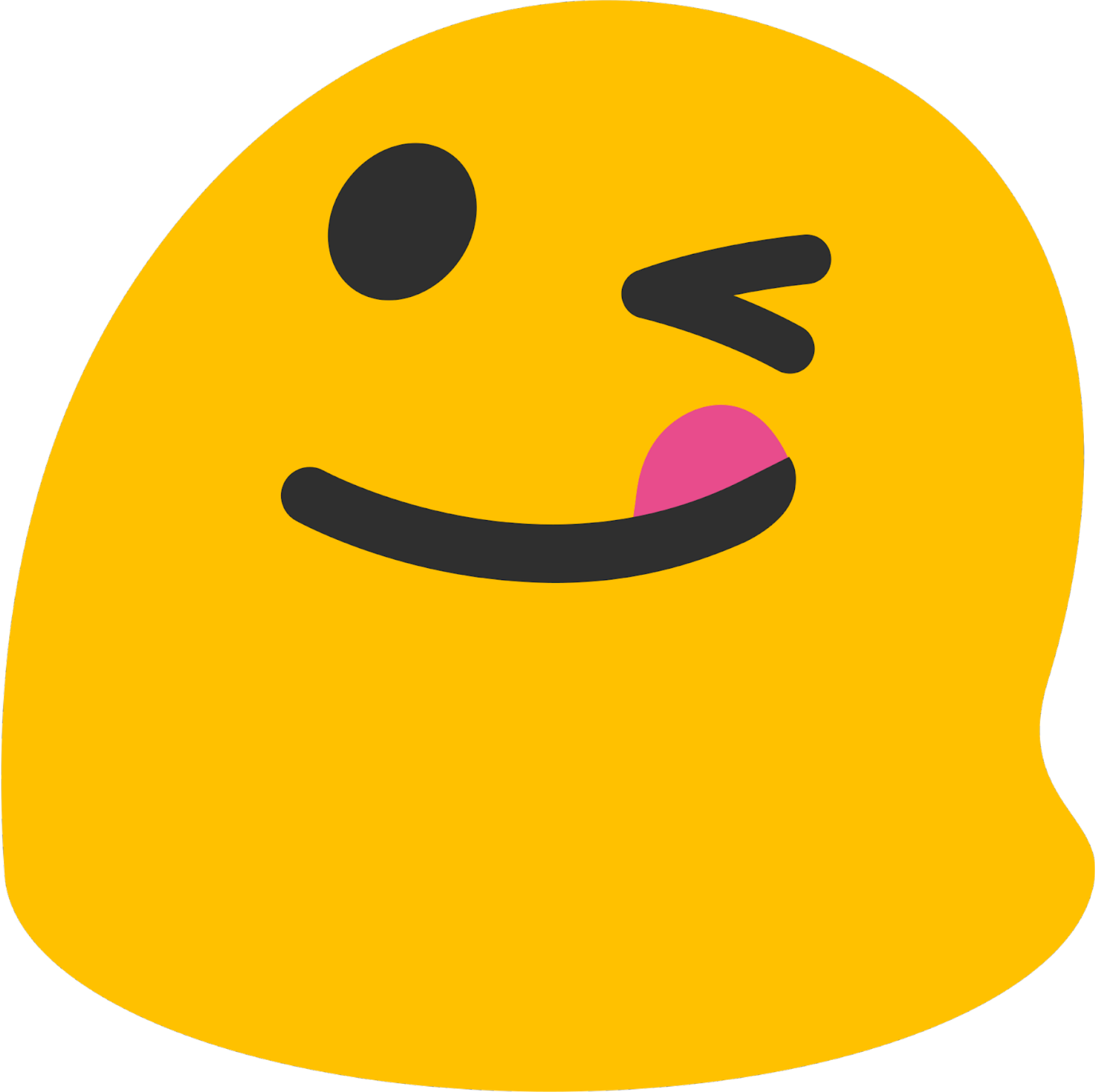}\hspace{0.5em}}
\newcommand*\crying{\includegraphics[width=1em]{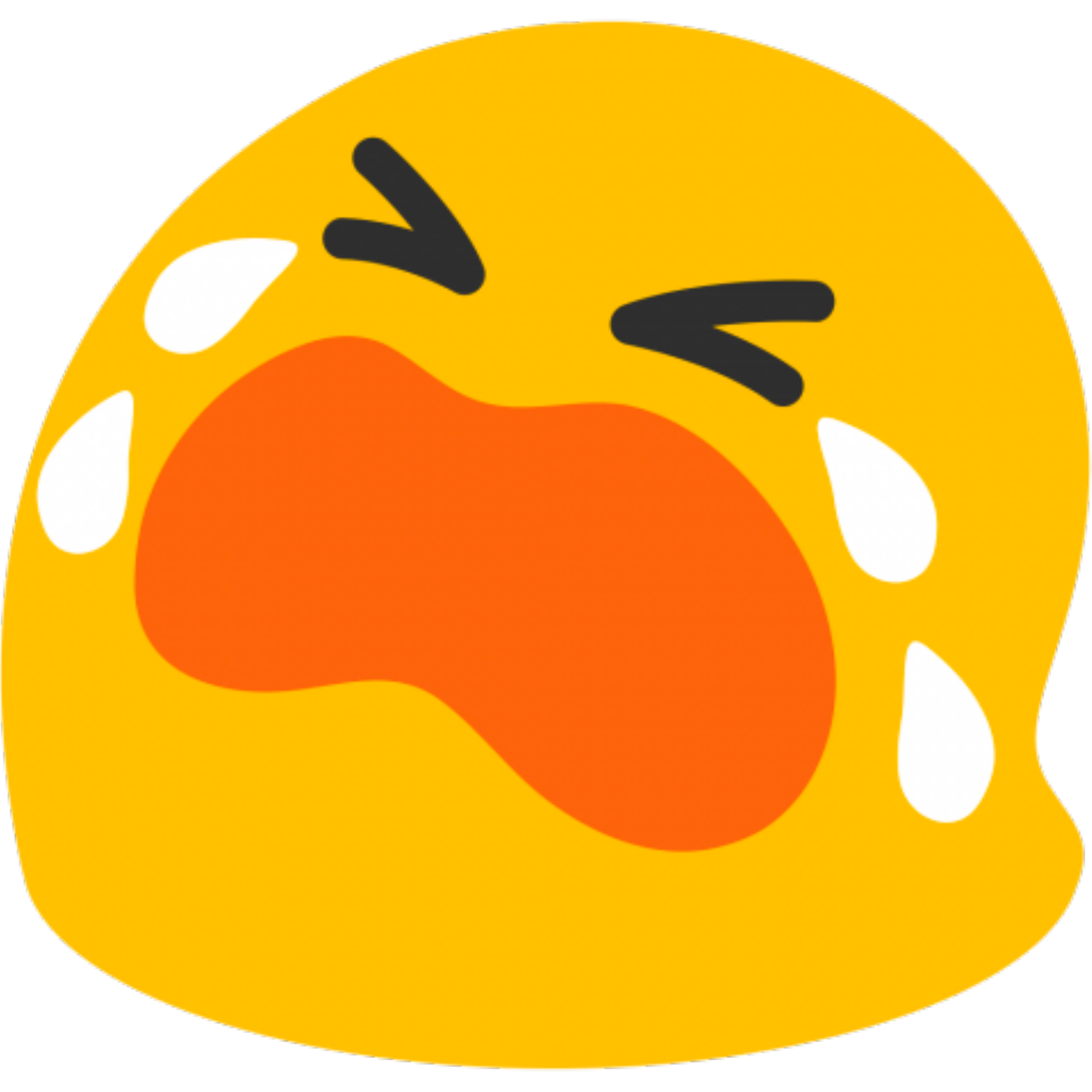}\hspace{0.5em}}

\title{\emph{Does Your Model Classify Entities Reasonably?} \\ Diagnosing and Mitigating Spurious Correlations in Entity Typing}

\author{Nan Xu, Fei Wang, Bangzheng Li, Mingtao Dong, Muhao Chen\\
Department of Computer Science \& Information Sciences Institute\\
University of Southern California\\
\texttt{\{nanx,fwang598,bangzhen,mingtaod,muhaoche\}@usc.edu}
}

\begin{document}
\maketitle
\begin{abstract}
Entity typing aims at predicting one or more words that describe the type(s) of a specific mention in a sentence.
Due to shortcuts from surface patterns to annotated entity labels and biased training, 
existing entity typing models are subject to the problem of \emph{spurious correlations}.
To comprehensively investigate the faithfulness and reliability of entity typing methods, 
we first systematically define distinct kinds of model biases that are reflected mainly from spurious correlations. 
Particularly, we identify six types of existing model biases, 
including \emph{mention-context} bias, \emph{lexical overlapping} bias, \emph{named entity} bias, \emph{pronoun} bias, \emph{dependency} bias, and \emph{overgeneralization} bias.
To mitigate model biases, we then introduce a counterfactual data augmentation method.
By augmenting the original training set with their debiased counterparts, models are forced to fully comprehend sentences and discover the fundamental cues for entity typing, rather than relying on spurious correlations for shortcuts.
Experimental results on the \emph{UFET} dataset show our counterfactual data augmentation approach helps improve generalization of different entity typing models with consistently better performance on both the original and 
debiased 
test sets\footnote{Code and resources are available at \url{https://github.com/luka-group/DiagnoseET}.}.
\end{abstract}
\section{Introduction}
\begin{figure}[t!]
\centering
\includegraphics[width=\linewidth]{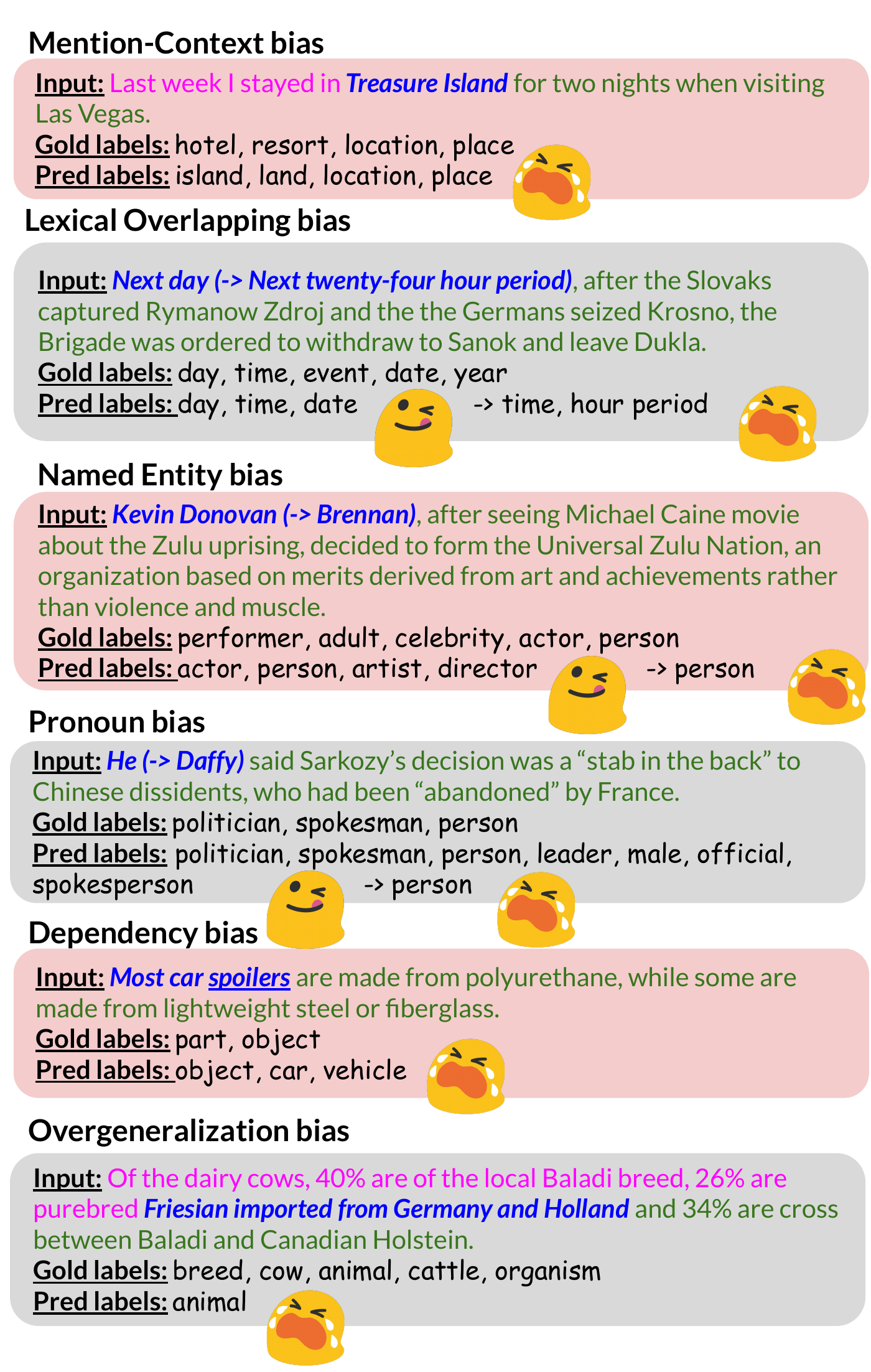}
\vspace{-1em}
\caption{Examples demonstrating spurious correlations exploited by one of the SOTA entity typing models \emph{MLMET}. \textcolor{magenta}{Left context} is in magenta, \textcolor{blue}{\emph{entity mention}} in italic blue,  \textcolor{green}{right context} in green. Perturbations upon mentions and new predictions start from $\textcolor{blue}{\rightarrow}/\rightarrow$. \smiley implies good predictions by exploiting spurious correlations and \crying indicates bad predictions when spurious correlations no longer exist. \emph{MLMET} falsely relies on the entity name to give ``island'' predictions for a hotel mention, incorrectly infers types of the dependent ``car'' rather than the headword ``spoiler'', and gives only the coarse label ``animal'' with more fine-grained missing.}
\label{fig:spurious_correlations}
\vspace{-1em}
\end{figure}


Given a sentence with an entity mention, the \emph{entity typing} task aims at predicting one or more 
words or phrases that describe the type(s) of that specific mention~\cite{ling2012fine,gillick2014context,choi-etal-2018-ultra}. 
This task essentially supports the structural perception of unstructured text \cite{distiawan2019neural}, being an important step for natural language understanding (NLU).
More specifically, entity typing has a broad impact on various NLP tasks that depend on type understanding, including coreference resolution~\cite{onoe-durrett-2020-interpretable}, entity linking 
\cite{hou-etal-2020-improving,tianran-etal-2021-improving}, entity disambiguation 
\cite{onoe-durrett-2020-interpretable}, event detection 
\cite{le-nguyen-2021-fine} and relation extraction~\cite{zhou2021improved}.

To tackle the task,
literature has developed various predictive methods to capture the association between the contextualized entity mention representation and the type label.
For instance, a number of prior studies approach the problem as multi-class classification based on distinct ways of representing the entity mentioning sentences~\cite{yogatama-etal-2015-embedding,ren2016label,xu-barbosa-2018-neural,dai-etal-2021-ultra}. Other studies formulate the problem as structured prediction and leverage structural  representations such as box embeddings~\cite{onoe-etal-2021-modeling} and causal chains~\cite{liu-etal-2021-fine} to model the dependency of type labels.
However, due to shortcuts from surface patterns to annotated entity labels and biased training, 
existing entity typing models are subject to the problem of \emph{spurious correlation}~\cite{wang-culotta-2020-identifying,wang2021identifying,branco-etal-2021-shortcutted}. 
For example, given a sentence ``Last week I stayed in \emph{Treasure Island} for two nights when visiting Las Vegas.'', a SOTA model like \emph{MLMET} \cite{dai-etal-2021-ultra} may overly rely on the entity name and falsely type \emph{Treasure Island} as an \emph{island}, while ignoring the sentential context that indicates this entity as a \emph{resort} or a \emph{hotel}. 
For morphologically rich mentions with multiple noun words such as ``most car spoilers'', entity models may fail to understand its syntactic structure and miss the target entity from the actual head-dependent relationship, leading to predictions describing the dependent car (\emph{car}, \emph{vehicle}) rather than the head spoilers (\emph{part}).
Such spurious clues
can cause the models to give unfaithful entity typing and further harm the machine's understanding of the entity mentioning text.

To comprehensively investigate the faithfulness and reliability of entity typing methods,
the \emph{first} contribution of this paper is to systematically define distinct kinds of model biases that are reflected mainly from spurious correlations. Particularly, we identify the following six types of existing 
model biases, for which examples are illustrated in \Cref{fig:spurious_correlations}. 
Those biases include \emph{mention-context} biases, \emph{lexical overlapping} biases, \emph{named entity} biases, \emph{pronoun} biases, \emph{dependency structure} biases and \emph{overgeneralization} biases.
We provide a prompt-based method to identify instances posing those biases to the typing model.
In the meantime, we illustrate that common existence of these types of biased instances causes it hard to evaluate whether a model is faithfully comprehending the entire context to infer the type, or trivially leveraging surface forms or distributional cues to guess the type.

We introduce a counterfactual data augmentation \cite{zmigrod-etal-2019-counterfactual} method for debiasing entity typing,
as the \emph{second} contribution of this paper. Given biased features, we reformulate entity typing as a type-querying cloze test and leverage a pre-trained language model (PLM) to fill in the blank. By augmenting the original training set with their debiased counterparts, models are forced to fully comprehend the sentence and discover the fundamental cues for entity typing, rather than rely on spurious correlations for shortcuts. Compared with existing debiasing approaches such as product of experts~\cite{he-etal-2019-unlearn}, focal loss~\cite{karimi-mahabadi-etal-2020-end}, contrastive learning~\cite{zhou-etal-2021-contrastive} and counterfactual inference~\cite{qian-etal-2021-counterfactual}, our counterfactual data augmentation approach helps improve generalization of all studied models with consistently better performance on both original \emph{UFET}~\cite{choi-etal-2018-ultra} and 
debiased 
test sets.


\section{Method}
In this section, we start with the problem definition (\Cref{sec:problem_definition}) and then categorize and diagnose the spurious correlations 
causing shortcut predictions by the typing model (\Cref{sec:spurious_correlation_definition_diagnoses}). 
Lastly, we propose a counterfactual data augmentation approach to mitigate the identified spurious correlations, 
as well as several alternative techniques that apply
(\Cref{sec:mitigating_spurious_correlations}).
\subsection{Problem Definition}\label{sec:problem_definition}
Given a sentence $s$ with an entity mention $e\in s$, the \emph{entity typing} task aims at predicting one or more words or phrases $T$ from the label space $L$ that describe the type(s) of $e$.

By nature, the inference of type $T$ should be context-dependent.
Take the first sample demonstrated in~\Cref{fig:spurious_correlations} as an instance: in ``Last week I stayed in \underline{\textbf{\emph{Treasure Island}}} for two nights when visiting Las Vegas,'' Treasure Island should be typed as \emph{hotel} and \emph{resort}, rather than \emph{island} or \emph{land} by 
trivially considering the surface of mention phrase.


\subsection{Spurious Correlations Diagnoses}\label{sec:spurious_correlation_definition_diagnoses}
\begin{table*}[t!]
\setlength{\tabcolsep}{1pt}
\resizebox{\textwidth}{!}{%
\begin{tabular}{@{}ll@{}}
\toprule
PLM Prompts & Entity Typing Instances                                   \\ \midrule\midrule
\begin{tabular}[t]{@{}l}
\textbf{Mention-Context}: \\
\hypertarget{p_mention_context}{\colorbox{lightgray}{$\mathbb{\bf\textcolor{red}{Prompt\:I}}$: \emph{<Mention> is a type of <mask>.}}}\\
\hypertarget{s1}{\textbf{\textcolor{red}{S1}}}: \underline{\textbf{\emph {fire}}} is a type of \emph{\textless{}mask\textgreater{}}.\\ \textbf{\emph{RoBERTa}}: \emph{energy, heat, explosion, fire, gas}\\\hypertarget{s2}{\textbf{\textcolor{red}{S2}}}: \underline{\textbf{\emph{the war}}} is a type of \emph{<mask>}.\\\textbf{True labels}: \emph{war, battle, conflict}\\\textbf{\emph{RoBERTa}}: \emph{war, battle, conflict, violence, warfare}\end{tabular}           & \begin{tabular}[t]{@{}l}\hypertarget{t1}{\textbf{\textcolor{blue}{T1}}}: A teacher who survived the shooting said he would never forgive\\the police for taking an hour to arrive after the gunman opened \underline{\textbf{\emph {fire}}}. \\ \textbf{True labels}: \emph{injury, shooting, event, violence}\\ \textbf{\emph{MLMET}}: \emph{event, object, fire} (F1: 0.285)\\ \cmidrule(r){1-1} \hypertarget{t2}{\textbf{\textcolor{blue}{T2}}}: \underline{\textbf{\emph {fire}}}\\ \textbf{\emph{MLMET}}: \emph{object, fire} (F1: 0.0)\end{tabular}
\\\midrule\midrule
\begin{tabular}[t]{@{}l}
\textbf{Lexical Overlapping}:\\
\hypertarget{p_lexical_overlapping}{\colorbox{lightgray}{$\mathbb{\bf\textcolor{red}{Prompt\:II}}$: \emph{<Left Context> <New Mention with Word Substitution> }}}\\\colorbox{lightgray}{\emph{<Right Context>. <New Mention with Word Substitution> is a type of}}\\\colorbox{lightgray}{\emph{<mask>.}}\\
\hypertarget{s3}{\textbf{\textcolor{red}{S3}}}: \underline{\textbf{\emph {Next twenty-four hour period}}}, after the Slovaks captured ...
\underline{\textbf{\emph {Next}}}\\\underline{\textbf{\emph{ twenty-four hour period}}} is a type of \emph{\textless{}mask\textgreater{}}.\\ \textbf{\emph{RoBERTa}}: \emph{confusion, retreat, ambush, battle, timeline}\end{tabular} & \begin{tabular}[t]{@{}l}\hypertarget{t3}{\textbf{\textcolor{blue}{T3}}}: \underline{\textbf{\emph {Next day}}}, after the Slovaks captured Zdroj the Brigade was\\ordered to withdraw to Sanok.\\ \textbf{True labels}: \emph{day, time, event, date, year}\\ \textbf{\emph{MLMET}}: \emph{day, time, date} (F1: 0.749)\\ \cmidrule(r){1-1} \hypertarget{t4}{\textbf{\textcolor{green}{T4}}}: \underline{\textbf{\emph {Next twenty-four hour period}}}, after the Slovaks captured Zdroj,\\the Brigade was ordered to withdraw to Sanok.\\ \textbf{\emph{MLMET}}: \emph{time, hour, period} (F1: 0.25)\end{tabular}\\
\midrule\midrule
\begin{tabular}[t]{@{}l}
\textbf{Named Entity}:\\
\hypertarget{p_named_entity}{\colorbox{lightgray}{$\mathbb{\bf\textcolor{red}{Prompt\:III}}$: \emph{The <Attribute> <Named Entity> is a type of <mask>.}}}\\
\hypertarget{s4}{\textbf{\textcolor{red}{S4}}}: The person \underline{\textbf{\emph {Benjamin Netanyahu}}} is a type of \emph{\textless{}mask\textgreater{}}.\\ \textbf{\emph{RoBERTa}}: \emph{politician, person, character, personality, man}
\\\hypertarget{s5}{\textbf{\textcolor{red}{S5}}}: The person \underline{\textbf{\emph{Jintara Poonlarp}}} is a type of \emph{\textless{}mask\textgreater{}}.\\ \textbf{\emph{RoBERTa}}: \emph{person, human, woman, personality, character}\end{tabular} 
& \begin{tabular}[t]{@{}l}\hypertarget{t5}{\textbf{\textcolor{blue}{T5}}}: \underline{\textbf{\emph{Benjamin Netanyahu}}} asserted that Amin al-Husseini had been\\one of the masterminds of the Holocaust.\\
\textbf{True labels}: \emph{politician, leader, person}\\ \textbf{\emph{MLMET}}: \emph{politician, leader, person} (F1: 1.0)\\ \cmidrule(r){1-1}\hypertarget{t6}{\textbf{\textcolor{green}{T6}}}: \underline{\textbf{\emph{Jintara Poonlarp}}} asserted that Amin al-Husseini had been one of\\ the masterminds of the Holocaust .\\ \textbf{\emph{MLMET}}: \emph{person, scholar, writer} (F1: 0.333)\end{tabular} \\
\midrule\midrule
\begin{tabular}[t]{@{}l}
\textbf{Pronoun}:\\
\hypertarget{p_pronoun}{\colorbox{lightgray}{$\mathbb{\bf\textcolor{red}{Prompt\:IV}}$: \emph{<Left Context> <Person Name> <Right Context>. }}}\\ \colorbox{lightgray}{\emph{<Person Name> is a type of <mask>.}}\\
\hypertarget{s6}{\textbf{\textcolor{red}{S6}}}: \underline{\textbf{\emph{Judith}}} other film credits include `` The Four Feathers,'' ``  Dr. T \&\\the Women '' and `` 200 Cigarettes.'' \underline{\textbf{\emph {Judith}}}  is a type of \emph{\textless{}mask\textgreater{}}.\\ \textbf{\emph{RoBERTa}}: \emph{bird, cat, vampire, rabbit, dog}\end{tabular}                                                                                                                                    & \begin{tabular}[t]{@{}l}\hypertarget{t7}{\textbf{\textcolor{blue}{T7}}}: \underline{\textbf{\emph{Her}}} other film credits include `` The Four Feathers,'' ``  Dr. T \&\\the Women '' and `` 200 Cigarettes.''\\ \textbf{Truths}: \emph{woman, performer, adult, female, entertainer, person, actress} \\                 \textbf{\emph{MLMET}}: \emph{woman, female, actress, person, artist} (F1: 0.666)\\ \cmidrule(r){1-1} \hypertarget{t8}{\textbf{\textcolor{green}{T8}}}: \underline{\textbf{\emph{Judith}}} other film credits include `` The Four Feathers,'' ``  Dr. T\\\& the Women '' and `` 200 Cigarettes.''\\ \textbf{\emph{MLMET}}: \emph{person} (F1: 0.25)\end{tabular}\\ \bottomrule
\end{tabular}%
}
\vspace{-0.5em}
\caption{Entity typing instances with content-based biases recognized by RoBERTa-large~\cite{liu2019roberta}. To reflect the shortcuts exploited by entity typing models (\Cref{sec:spurious_correlation_definition_diagnoses}), we list the sentences, labels and predictions from one of the SOTA models \emph{MLMET} in \textcolor{blue}{T1}, \textcolor{blue}{T2}, \textcolor{blue}{T3}, \textcolor{blue}{T5} and \textcolor{blue}{T7}. To identify biased instances (\Cref{sec:spurious_correlation_definition_diagnoses}), we show the constructed masked fill-in task to query the PLM with mention types from \textcolor{red}{S1} to \textcolor{red}{S6}. To mitigate spurious correlations (\Cref{sec:mitigating_spurious_correlations}), we show the proposed counterfactual data augmentation where the shortcuts disappear and the model fails in \textcolor{green}{T4}, \textcolor{green}{T6} and \textcolor{green}{T8}. We underline the \underline{\textbf{\emph{mention span}}} in italic boldface and record the macro F1 score for each prediction.}
\label{tab:bias_example_I}
\vspace{-0.5em}
\end{table*}

We systematically define six types of typical model biases caused by spurious correlations in entity typing models. For each bias, we qualitatively inspect its existence and the corresponding spurious correlations used by a SOTA entity typing model on sampled instances with bias features. 
Following \citet{poerner-etal-2020-e}, we prompt a PLM, \emph{RoBERTa-large}~\cite{liu2019roberta}, to identify potential biasing samples with either detected surface patterns 
or facts 
captured during training. To do so, 
we reformulate entity typing as a type-querying cloze task and perform the analysis as follows.

\textbf{\emph{1) Mention-Context Bias}}:
Semantically rich entity mentions 
may encourage the model to overly associate the mention surface with the type 
without considering the key information stated in contexts. 
An example is accordingly shown in \hyperlink{t1}{T1} of~\Cref{tab:bias_example_I}, where \emph{MLMET} predicts types that correspond to the case where ``fire'' is regarded as \emph{burning} instead of \emph{gun shooting}. Evidently, this is due to not effectively capturing the clues in the context such as ``shooting'' and ``gunman''.
This is further illustrated by the counterfactual example \hyperlink{t2}{T2}, where the model predicts almost the same labels when seeing ``fire'' without a context.

To identify potential instances with the mention-context bias, we query the PLM to infer the entity types based only on the mention 
with the template shown in \hyperlink{p_mention_context}{Prompt I} (\Cref{tab:bias_example_I}).
Therefore, samples where the PLM can accurately predict without the context information are regarded as biased. 
Entity typing models can easily achieve good performance on those biased samples by leveraging spurious correlations between their mention surface and types, as shown in \hyperlink{s2}{S2} from~\Cref{tab:bias_example_I}. 

\begin{table*}[t!]
\resizebox{\textwidth}{!}{%
\begin{tabular}{@{}lll@{}}
\toprule
Bias Type                                                                                                                                                                                                         & Analyses                                                                                                                                                                                                                                                                                         & Entity Typing Instances\\ \midrule\midrule
\begin{tabular}[t]{@{}ll}\textbf{Dependency}:\\ Models fail to capture the \\ syntactic structure and \\ make type predictions \\ focusing on other \\ components rather \\ than the dependency \\ headword\end{tabular} & \begin{tabular}[t]{@{}ll}  \hypertarget{s6}{\textbf{\textcolor{green}{S6} Dependency Parsing}}: \\ 
\underline{
token | text | head text}\\ 1st\quad\quad whale  \quad anatomy\\ 2nd \quad  anatomy\quad anatomy\end{tabular}                                                                                           & \begin{tabular}[t]{@{}ll}\hypertarget{t9}{\textbf{\textcolor{blue}{T9}}}: Dubois contributed an article on \underline{\textbf{\emph{whale anatomy}}} to a book by the Dutch zoologist ,  Max \\Wsubjecteber , and , inspired by the fresh discovery of new Neanderthal fossils at  the Belgian\\ town of Spy , he spent his vacation fossil hunting in the vicinity of his birthplace .\\ \textbf{True labels}: \emph{subject, topic}\\ \textbf{\emph{MLMET}}: \emph{object, animal} (F1: .0)\\\cmidrule(r){1-1} \hypertarget{t10}{\textbf{\textcolor{green}{T10}}}: Dubois contributed an article on \underline{\textbf{\emph{anatomy}}} to a book by the Dutch zoologist ,  Max \\Wsubjecteber , and , inspired by the fresh discovery of new Neanderthal fossils at the  Belgian\\ town of Spy, he spent his vacation fossil hunting in the vicinity of his birthplace .\\ \textbf{\emph{MLMET}}: \emph{object, concept, subject} (F1: .4)\end{tabular} \\\midrule\midrule
\begin{tabular}[t]{@{}ll}\textbf{Overgeneralization}:\\ Models suffer from \\ biased training \\ due to extreme class \\ imbalance\end{tabular}                                                                          & \begin{tabular}[t]{@{}ll}\textbf{\textcolor{blue}{Statistics}}:\\\textbf{Counts of Coarse Types}:\\\emph{person}: 824, \emph{event}: 181\\\textbf{Counts of Ultra-fine Types}:\\ \emph{concept}: 68, \emph{activity}: 23,\\ \emph{trouble}: 8, \emph{difficulty}: 6,\\ \emph{problem}: 5, \emph{misconduct}: 1,\\\emph{use}: 1, \emph{abuse}: 0\\ \emph{behavior}: 0, \emph{wrongdoing}: 0\end{tabular} & \begin{tabular}[t]{@{}ll}\hypertarget{t11}{\textbf{\textcolor{blue}{T11}}}: \underline{\textbf{\emph{Many nineteenth-century individualist anarchists ,}}} including Benjamin Tucker , rejected\\ the anarcho-capitalist Lockean position in favour of the anarchist position of '' occupancy and \\use '' -LRB- or '' possession '' , to use Proudhon 's term -RRB- , particularly in land .\\ \textbf{True labels}: \emph{person}\\ \textbf{\emph{MLMET}}: \emph{person} (F1: 1.0)\\\cmidrule(r){1-1}
\hypertarget{t12}{\textbf{\textcolor{blue}{T12}}}: In a letter , the exchange said its investigations had turned up `` no evidence  of \underline{\textbf{\emph{abusive}}}\\ \underline{\textbf{\emph{behavior}}} . ''\\ \textbf{True labels}: \emph{behavior, wrongdoing, difficulty, misconduct, trouble, use, activity, problem, } \\ \emph{concept, abuse}\\ \textbf{\emph{MLMET}}: \emph{behavior, event} (F1: .166) \\\cmidrule(r){1-1}
\hypertarget{t13}{\textbf{\textcolor{red}{T13}}}: <NULL Input>\\
\textbf{\emph{MLMET}}:\emph{person }(prob=0.992)\end{tabular}
\\ \bottomrule
\end{tabular}%
}
\vspace{-0.5em}
\caption{Entity typing instances from \emph{UFET} test set with biases detected based on statistical analyses. To discover shortcuts utilized by entity typing models (\Cref{sec:spurious_correlation_definition_diagnoses}), we show one \emph{Dependency} bias instance  where the model fails to locate the target entity in the mention (\textcolor{blue}{T9}) and two \emph{Overgeneralization} bias instances: \textcolor{blue}{T11} annotated by coarse types and \textcolor{blue}{T12} annotated by ultra-fine types. To quantify the overgeneralization bias (\Cref{sec:spurious_correlation_definition_diagnoses}), we query the typing model with an empty sentence in \textcolor{red}{T13}. To mitigate spurious correlations (\Cref{sec:mitigating_spurious_correlations}), we do dependency parsing to distinguish headwords from dependents in \textcolor{green}{S6} and truncate the mention with only the headword preserved as \textcolor{green}{T10} to help address dependency bias.}
\label{tab:bias_example_II}
\vspace{-0.5em}
\end{table*}

\textbf{\emph{2) Lexical Overlapping Bias}}: 
Type labels that have lexical overlaps with the entity mention can also become prediction shortcuts.
As shown in \hyperlink{t3}{\textbf{T3}} from~\Cref{tab:bias_example_I}: 
labeling mention ``next day'' with the type \emph{day} and additional relevant types leads to the F1 up to $0.749$.
We observe a considerable amount of similar examples,
e.g., typing the mention ``eye shields'' as \emph{shield}, ``the Doha negotiations'' as \emph{negotiation}, 
etc. 
The highly overlapped mention words and type labels
make it difficult to evaluate whether the model makes predictions based on  content comprehension or simply 
lexical similarities.

We substitute the overlapping mention words 
with  semantically similar words and ask the PLM to infer the entity types on such perturbed instances (details introduced in~\Cref{sec:mitigating_spurious_correlations}) 
by prompting with the template  \hyperlink{p_lexical_overlapping}{Prompt II} (\Cref{tab:bias_example_I}).
We consider instances have lexical overlapping biases when the PLM performs poorly after 
the overlapped mention words are substituted, as shown in 
\hyperlink{s3}{S3} of~\Cref{tab:bias_example_I}.

\textbf{\emph{3) Named Entity Bias}}: 
On cases where mentions refer to high-reporting entities in corpora, 
models may be trained to ignore the context but directly predict labels that co-occur frequently with those entities.
We show a concrete instance to type a person named entity in \hyperlink{t5}{\textbf{T5}} of~\Cref{tab:bias_example_I}. 
The mention \emph{Benjamin Netanyahu}, known as Israeli former prime minister, 
is normally annotated with \emph{politician, leader} and \emph{authority}.
After observing popular named entities and their common annotations during training, models are able to predict 
their common types,
making it hard to evaluate models' capabilities to infer context-sensitive labels.

As illustrated in \hyperlink{p_named_entity}{Prompt III} (\Cref{tab:bias_example_I}), we prompt the PLM to type the named entity when only the name and its general attribute is given, e.g., \emph{the geopolitical area India} or \emph{the organization Apple}, 
etc. 
We regard instances to have the named entity bias when the PLM accurately infers the mention types relying on prior knowledge of named entities.
In \Cref{tab:bias_example_I}, we show one instance with the mention containing \emph{Benjamin Netanyahu} in \hyperlink{s4}{S4}, and the Thai pop music singer -- \emph{Jintara Poonlarp} in \hyperlink{s5}{S5}\footnote{Both represent celebrities reported by their own Wikipedia pages and thousands of news articles, hence are very likely to be covered by the pre-training corpora of \emph{RoBERTa}.}. 
Based on types related to \emph{Benjamin}'s political role in \hyperlink{s4}{S4} and general types for \emph{Jintara} in \hyperlink{s5}{S5}, we consider instances to type mentions including \emph{Benjamin} as biased while those with \emph{Jintara} as unbiased.

\textbf{\emph{4) Pronoun Bias}}: 
Compared with diverse person names, pronouns show up much more frequently to help make sentences smoother and clearer. 
Therefore, models are subject to biased training to type pronouns well, but lose the ability to type based on diverse real names.
To type the pronoun \emph{her} in \hyperlink{t7}{\textbf{T7}} of \Cref{tab:bias_example_I},
the entity typing model can successfully infer general types \emph{woman, female} as well as the context-sensitive type \emph{actress}. To obtain high generalization, we expect models to infer types correctly for both pronouns and their referred names. 

We substitute the gender pronoun with a random person name of the same gender (details introduced in~\Cref{sec:mitigating_spurious_correlations}) and ask the PLM to infer the types with \hyperlink{p_pronoun}{Prompt IV} (\Cref{tab:bias_example_I}).
We consider samples to have the pronoun bias when the PLM fails to capture the majority of types after the name substitution, as shown in 
 \hyperlink{s6}{S6} of~\Cref{tab:bias_example_I}.

\textbf{\emph{5) Dependency Bias}}: 
It is observed that the mention's headwords explicitly match the mention to its types
~\cite{choi-etal-2018-ultra}. However, models may fail to capture the syntactic structure with predictions focusing on dependents instead of headwords.
We show an instance with inappropriate focus among mention words in \hyperlink{t9}{\textbf{T9}} of~\Cref{tab:bias_example_II}.
Without understanding the mention's syntactic structure, entity typing models may make predictions that are irrelevant to the actual entity. 

Since knowledge about mention structures is beneficial for typing complex multi-word mentions, we mitigate the bias by data augmentation to improve model learning (details introduced in~\Cref{sec:mitigating_spurious_correlations}), rather than identify whether the bias exists or not.

\textbf{\emph{6) Overgeneralization Bias}}: 
When training with disproportional distributed labels, 
frequent labels are more likely to be predicted compared with rare ones. Entity typing datasets are naturally imbalanced~\cite{gillick2014context,choi-etal-2018-ultra}.
We show two instances annotated by coarse- and fine-grained labels in \hyperlink{t11}{\textbf{T11}} and \hyperlink{t12}{\textbf{T12}} of~\Cref{tab:bias_example_II}: the model can easily predict the coarse-grained label \emph{person}
to describe ``anarchist'', but fails to infer less frequent but more concrete labels such as \emph{misconduct} and \emph{wrongdoing} to type \emph{behavior}. Models ought to type entities by reasoning on mentions and contexts, rather than trivially fitting the label distribution.

As shown in \hyperlink{t13}{T13} of~\Cref{tab:bias_example_II},
we craft a special instance -- an empty sentence, with which the uniform distribution over all types is expected from models free of overgeneralization bias. We then compute its disparity with the model's actual probability distribution:
the higher/lower probability predicted on popular/rare types, the more biased the model on the label distribution. 

\stitle{Discussion}
The prior defined six biases are not mutually exclusive. 
We discuss some possible 
mixtures of concurrent biases as follows:

\emph{ Mention-Context} and \emph{Lexical Overlapping Bias}: the model falsely types the mention ``Treasure Island'' as \emph{island}, without understanding the context talking about the holiday accommodation. Another possible reason that the mention far outweighs the context might be the high word similarity between  mention word ``Island'' and  type word ``island''. 

\emph{ Dependency} and \emph{Lexical Overlapping Bias}: \emph{MLMET} incorrectly makes the prediction \emph{car}  for the mention ``most car spoilers'' without distinguishing important headwords from less important dependent words. Another reasonable explanation for emphasizing on the dependent rather than the headword is its perfect lexical match with the type set, where ``car'' is a relatively popular label but no type has high word similarity with ``spoilers''. 
To diagnose and mitigate all spurious correlations the entity typing model may take advantage of, we disentangle the multiple biases on a single instance by \emph{analyzing each bias individually without considering their mutual interactions}. 

\subsection{Mitigating Spurious Correlations}\label{sec:mitigating_spurious_correlations}
Models exploiting spurious correlations lack the required reasoning capability, leading to unfaithful typing and 
harmed out-of-distribution generalization when bias features observed during training do not hold.
Therefore, we propose to mitigate spurious correlations 
from the counterfactual data augmentation perspective:
for each instance recognized with specific bias features, we automatically craft its debiased counterpart and train entity typing models with both samples. Whenever the model prefers to exploit 
biasing features, it will fail on newly crafted debiased instances and actively look for 
more robust features: 
understanding and reasoning on the sentence rather than exploiting spurious correlations. 
Considering the characteristic textual patterns from different biases, we propose the following distinct strategies to craft debiased instances 
for four types of biases (with examples explained in~\Cref{sec:mitigate_examples}).  
Note that although we can hardly craft a new instance free of mention-context bias or overgeneralization bias, 
we can choose to leverage the alternative debiasing techniques introduced in later parts of this section for these two biases.

\stitle{Counterfactual Augmentation}
On instances diagnosed with lexical overlapping biases, 
we perform word substitutions in two steps to substitute mention words lexically similar to type labels with original semantics preserved. 
To do so, we identify the sense of type words in mentions using an off-the-shelf word sense disambiguation model \cite{barba-etal-2021-esc} and substitute them with their WordNet synonyms.
We consider perturbed sentences with poor performance from the PLM 
as the \emph{counterfactual augmented} instances free from lexical overlapping bias, to prohibit the entity typing model from exploiting spurious correlations (\hyperlink{t4}{T4} of~\Cref{tab:bias_example_I}). 

For instances with the named entity bias,
we augment by performing named entity substitution according to the following criteria. \emph{1) validity}: substituted entities should 
have the same general type as the original ones\footnote{We consider the 12 NER types including person, geopolitical area, location, organization, group, date, facility, work of art, ordinal number, event, product, and time.}, e.g., the geopolitical area ``India'' can be replaced by  ``London''; 
\emph{2) debiased}: models training on large corpora should not possess comprehensive knowledge of the new named entities.
Basically, we leverage an off-the-shelf NER model \cite{ushio-camacho-collados-2021-ner} to identify and classify named entities 
into general NER types provided by this model, and then divide the entities into \textbf{informative} and \textbf{non-informative} group based on the prompt-based typing performance by the PLM. We then substitute informative named entities with non-informative ones sharing the same NER type 
as the \emph{counterfactual augmented} instances (\hyperlink{t6}{T6} of~\Cref{tab:bias_example_I}).


For the   pronoun bias,
we craft new instances by concretizing pronoun mentions in two situations. 
If co-reference resolution \cite{toshniwal-etal-2021-generalization} detects the referred entity of the pronoun mention in the context, that entity is selected as the new mention.
Otherwise, the gender pronoun mention will be substituted with a randomly sampled masculine/feminine name from the NLTK corpus \cite{bird-2006-nltk}.
New sentences with the actual person names are considered \emph{counterfactual augmented} if the PLM fails to infer the person's type with contextual information given (\hyperlink{t7}{T7} of~\Cref{tab:bias_example_I}).


We further augment from instances where mentions have internal dependency structures to tackle the dependency bias.
First, we use a dependency parsing tool~\cite{Honnibal_spaCy_Industrial-strength_Natural_2020} 
to recognize the dependency parse tree of the mention.
On top of that, we truncate all other dependent words in the new mention to create the 
augmentation.
From associations between explicitly provided headwords and their matching labels, the models are encouraged to learn dependency structures for targeted entity typing and predict precisely when headwords and dependents are mixed in mentions (\hyperlink{t9}{T9} of~\Cref{tab:bias_example_II}).

Together with the new instances with headwords explicitly given, instances counterfactually augmented upon the entity typing training set is utilized to allow various entity typing models to learn to mitigate spurious correlations. Meanwhile, we leverage the counterfactual augmented instances derived from the test set for model evaluation.
\begin{table*}[t!]

\resizebox{\textwidth}{!}{%
\setlength{\tabcolsep}{3pt}
\begin{tabular}{@{}lcccccccc|cccc@{}}
\toprule
\multirow{2}{*}{Test Set} & \multicolumn{2}{c}{\begin{tabular}[c]{@{}c@{}}Mention-Context\\ 131/1085 ($\uparrow$)\end{tabular}}                              & \multicolumn{2}{c}{\begin{tabular}[c]{@{}c@{}}Lexical Overlapping\\ 475/3 ($\downarrow$)\end{tabular}}                               & \multicolumn{2}{c}{\begin{tabular}[c]{@{}c@{}}Named Entity\\ 36/500 ($\downarrow$)\end{tabular}}                                     & \multicolumn{2}{c|}{\begin{tabular}[c]{@{}c@{}}Pronoun\\ 881/20 ($\downarrow$)\end{tabular}}                                        & \multicolumn{2}{c}{\begin{tabular}[c]{@{}c@{}}Dependency\\ 280, 222/961\end{tabular}}                                & \multicolumn{2}{c}{\begin{tabular}[c]{@{}c@{}}Overgeneralization\\ 93/242\end{tabular}} \\\cmidrule(r){2-13}
                          & BiLSTM                                                   & MLMET                                                    & BiLSTM                                                    & MLMET                                                     & BiLSTM                                                    & MLMET                                                     & BiLSTM                                                    & MLMET                                                    & BiLSTM                                                    & MLMET                                                    & BiLSTM                                      & MLMET                                     \\\midrule
Biased                    & .385                                                    & .654                                                    & .510                                                     & .551                                                     & .504                                                     & .735                                                     & .494                                                     & .561                                                    & .152                                                     & .424                                                    & .466                                       & .427                                     \\
-Perturb.                 & \begin{tabular}[c]{@{}c@{}}.400\\ (3.8\%)\end{tabular}  & \begin{tabular}[c]{@{}c@{}}.654\\ (0.0\%)\end{tabular}  & \begin{tabular}[c]{@{}c@{}}.050\\ (-90.3\%)\end{tabular} & \begin{tabular}[c]{@{}c@{}}.327\\ (-40.8\%)\end{tabular} & \begin{tabular}[c]{@{}c@{}}.332\\ (-34.0\%)\end{tabular} & \begin{tabular}[c]{@{}c@{}}.600\\ (-18.4\%)\end{tabular} & \begin{tabular}[c]{@{}c@{}}.179\\ (-63.8\%)\end{tabular} & \begin{tabular}[c]{@{}c@{}}.525\\ (-6.4\%)\end{tabular} & \begin{tabular}[c]{@{}c@{}}.396\\ (160.5\%)\end{tabular} & \begin{tabular}[c]{@{}c@{}}.564\\ (32.9\%)\end{tabular} & .118                                       & .232                                     \\ \midrule
Unbiased                  & .265                                                    & .436                                                    & .392                                                     & .544                                                     & .316                                                     & .505                                                     & .366                                                     & .683                                                    &  -                                                         &   -                                                       &     -                                        &       -                                    \\
-Perturb.                & \begin{tabular}[c]{@{}c@{}}.253\\ (-4.7\%)\end{tabular} & \begin{tabular}[c]{@{}c@{}}.395\\ (-9.4\%)\end{tabular} & \begin{tabular}[c]{@{}c@{}}.167\\ (-57.5\%)\end{tabular} & \begin{tabular}[c]{@{}c@{}}.444\\ (-18.2\%)\end{tabular} & \begin{tabular}[c]{@{}c@{}}.372\\ (17.6\%)\end{tabular}  & \begin{tabular}[c]{@{}c@{}}.539\\ (6.9\%)\end{tabular}   & \begin{tabular}[c]{@{}c@{}}.130\\ (-64.6\%)\end{tabular} & \begin{tabular}[c]{@{}c@{}}.659\\ (-3.5\%)\end{tabular} & -                                                          & -                                                         &  -                                           &        -                                   \\ \bottomrule
\end{tabular}%
}
\vspace{-0.5em}
\caption{F1 scores of two representative entity typing models on UFET testing samples with(out) distinct biases and their perturbations: mention-only input for \emph{Mention-Context}, overlapped word substitution for \emph{Lexical Overlapping}, named entity substitution for \emph{Named Entity}, name substitution for \emph{Pronoun}. Below each bias, the number of model-agnostic biased and unbiased instances are listed and $\downarrow/\uparrow$ indicates expected performance from models leveraging spurious correlations after perturbing biased instances. Relative performance drop/increase after testing on their perturbations is recorded in brackets. For \emph{Dependency} bias, we show performance on 280 and 222 out of 961 test samples where the two models benefit from making predictions based on headwords and contexts respectively. For \emph{Overgeneralization} bias, we show performance on 93/242 samples annotated by purely coarse/ultra-fine types (values on different subsets hence incomparable). See results of all five models evaluated by full metrics in~\Cref{tab:ufet_full}.}
\label{tab:diagnosing}
\vspace{-0.5em}
\end{table*}

\stitle{
Alternative Debiasing Techniques} 
In addition to data augmentation, other applicable debiasing techniques can be used to resample or reweight original instances in training, or directly measure and deduct biases in inference.
A typical resampling technique is \emph{AFLite}~\cite{le2020adversarial} which drops samples predicted accurately by simple models such as fasttext~\cite{joulin-etal-2017-bag}.
Reweighting techniques typically train one or more models to proactively identify and up-weight underrepresented instances in the training process, which includes product of experts, debiased focal loss, learned-mixin and its variant learned-mixin+H~\cite{clark-etal-2019-dont,he-etal-2019-unlearn,karimi-mahabadi-etal-2020-end}.
On the other hand, counterfactual inference \cite{qian-etal-2021-counterfactual} measures prediction biases based on counterfactual examples (e.g. masking out the context for measuring mention-context biases, or giving empty inputs to measure overgeneralization biases \cite{wang2022should}),
and directly deducts the biases in inference.
In addition, contrastive learning~\cite{chen2021exploring,caron2021emerging,chen2021exploring} can be used to adopt a contrastive training loss \cite{caron2021emerging,chen2021exploring} to discourage the model from learning similar representations for full and bias features\footnote{Models are discouraged to learn similar representations between full features and bias features such as mention, named entity, or pronoun as input. They are also encouraged to learn similar representations between original instances and counterfactual augmented instances with overlapped word substitution, instances with headwords as new mentions.}. 
Next, we compare our approach with those techniques.

\section{Experiments}\label{sec:exp}

In this section, we start with describing the experimental setups (\Cref{sec:setup}). Next, we diagnose entity models to measure their reliance on spurious correlations 
(\Cref{sec:diagnosing}). We then compare our counterfactual data augmentation with other debiasing techniques for spurious correlation mitigation (\Cref{sec:mitigating}).
\subsection{Experimental Settings}\label{sec:setup}
We leverage the ultra-fine entity typing (\textbf{UFET}) dataset~\cite{choi-etal-2018-ultra} to evaluate entity typing models and apply different mitigation approaches 
either during training or as inference post-processing. 
UFET comes with $6K$ samples from crowdsourcing and $25.2M$ distant supervision samples. There are $10,331$ types in total, among which nine are general (e.g., person), $121$ are fine-grained (e.g., engineer), and $10,201$ are ultra-fine (e.g., flight engineer). We follow prior studies \cite{choi-etal-2018-ultra} to evaluate entity typing models with macro-averaged precision, recall and F1. 
We also study spurious correlations and effectiveness of the proposed debiasing approach on \textbf{OntoNotes}~\cite{gillick2014context}.
As results present similar observations,
we leave detailed analysis in~\Cref{sec:ontonotes}.

\stitle{Entity Typing Baselines} 
We diagnose the 
prediction biases and the effectiveness of distinct debiasing models based on following approaches:
1) \textbf{BiLSTM}~\cite{choi-etal-2018-ultra} concatenates the context representation learned by a bidirectional LSTM and the mention representation learned by a CNN to predict entity labels.
2) \textbf{LabelGCN}~\cite{xiong-etal-2019-imposing} introduces  graph propagation to encode global label co-occurrence statistics and their word-level similarities.
3) \textbf{LRN}~\cite{liu-etal-2021-fine} autoregressively generates entity labels from coarse to fine levels, 
 modeling the coarse-to-fine label dependency as causal chains.
4) \textbf{Box4Types}~\cite{onoe-etal-2021-modeling} proposes to embed concepts as d-dimensional hyper rectangles (boxes), so that hierarchies of types could be captured as topological relations of boxes.
5) \textbf{MLMET}~\cite{dai-etal-2021-ultra} augments training data by constructing mention-based input for BERT to predict context-dependent mention hypernyms for type labels.
Without loss of generality, we discuss results of two representative models, the earliest \emph{BiLSTM} training from scratch and the latest \emph{MLMET} finetuning on the PLM, for the sake of clarity in this section. As the observations on the other models are similar, we leave those results in~\Cref{sec:appendix}.

\subsection{Diagnosing Entity Typing Models}\label{sec:diagnosing}
In~\Cref{tab:diagnosing}, we report performance of entity typing models trained on UFET. The models are tested on original biased samples and their perturbed new instances to reflect exploited spurious correlations.
We conduct similar analyses on unbiased samples.

\textbf{\emph{1) Mention-Context Bias}}:
When perturbing the biased samples by only feeding their mentions to typing models, the performance of \emph{MLMET} keeps unchanged while the performance of \emph{BiLSTM} even improves by $3.8\%$. 
This disobeys the task goal of 
entity typing where types of the mentions should also depend on contexts,
and we suggest that 
samples with mention-context biases
are insufficient for a 
faithful evaluation of a reliable typing system.


\textbf{\emph{2) Lexical Overlapping Bias}}: 
After substituting label-overlapped mention words with semantically similar words,
performance of both models drops drastically especially on biased samples identified by the PLM. 
Compared with \emph{MLMET}, 
\emph{BiLSTM} has less parameter capacity and is more inclined to 
leverage lexical overlapping between mentions and type labels as the shortcut for typing. 

Compared with original biased instances, the perturbed instances with label-overlapped mention words replaced might look less natural or fluent. In \Cref{tab:word_perturbation}, we therefore substitute words from different parts of instance, and prove that performance degradation is caused by removed lexical overlapping bias rather than unnatural or dysfluent input.   
\begin{table}[t!]
\resizebox{\columnwidth}{!}{%
\begin{tabular}{@{}lll@{}}
\toprule
Perturbed Words            & BiLSTM   & MLMET    \\ \midrule
Label-overlapped Mention     & -63.84\% & -38.60\% \\
Non Label-overlapped Mention & -2.77\%  & -11.64\% \\
Context                      & 3.86\%   & -0.18\%  \\ \bottomrule
\end{tabular}%
}
\caption{F1 performance variation after perturbing words at different parts of instances from UFET test with their synonyms. Much higher performance drop after replacing label-overlapped mention words proves the degradation caused by removing label overlapping bias rather than potential reduced naturalness or fluency due to word substitution. }
\label{tab:word_perturbation}
\vspace{-1.1em}
\end{table}

\textbf{\emph{3) Named Entity Bias}}:
After replacing named entities 
to be less impacted from biased prior knowledge, 
performance of both studied models in~\Cref{tab:diagnosing} decreases considerably when encountering named entities, with which models struggle to capture spurious correlations with mention types. 
Interestingly, perturbing unbiased samples by utilizing named entities with bias provides shortcuts for prediction, leading to improved performance of both models.

\textbf{\emph{4) Pronoun Bias}}: 
With pronouns replaced by their referred entities in contexts or random masculine/feminine names otherwise, 
we observe serious performance degradation from both models,
which demonstrates their common weakness on typing more diverse and less frequent real names.

\textbf{\emph{5) Dependency Bias}}: 
With headwords directly exposed to entity typing models by dropping all other less important dependents,
performance from \emph{BiLSTM} on around $30\%$ of all testing samples with dependency structures gets improved dramatically, while \emph{MLMET} also predicts more precisely on $23\%$ of samples. 
Hereby, we confirm that existing entity models still suffer from extracting core components of given mentions for entity typing and appeal for more research efforts to address this problem. 


\textbf{\emph{6) Overgeneralization Bias}}:
Models are subject to making biased predictions towards popular types observed during training, which leads to contrastive performance on instances purely annotated by coarse and ultra-fine types, as shown in~\Cref{tab:diagnosing}.
This problem is exemplified in a case study in~\Cref{tab:overgeneralization}, where typing models are queried with an empty sentence. 
Compared with the uniform probability distribution expected from models free from overgeneralization bias, existing models are 
inclined to give much higher probabilities to coarse types such as \emph{person} and \emph{title}.
\begin{table}[t!]
\resizebox{\columnwidth}{!}{%
\begin{tabular}{@{}ll@{}}
\toprule
Model     & Top/Bottom Types (Prob.)                                                                                                                                       \\ \midrule
BiLSTM & \begin{tabular}[c]{@{}l@{}}person (.928), title (.437), concept (.104)...\\ ...vice squad (.000), adolescent (.000), supporter (.000)\end{tabular}     \\\midrule
LabelGCN  & \begin{tabular}[c]{@{}l@{}}concept (.349), increase (.249), case (.192)...\\ ...archipelago (.000), spiritual leader (.000), national park (.000)\end{tabular} \\\midrule
Box4Types & \begin{tabular}[c]{@{}l@{}}object (.626), person (.282), company (.231)...\\ ...dismissal (.000), trump (.000), small town (.000)\end{tabular}                 \\ \midrule
LRN       & \begin{tabular}[c]{@{}l@{}}person (.998), writer (.000), place (.000)...\\ ...chicken leg (.000), chicken wing (.000), chicken wire (.000)\end{tabular}        \\ \midrule
MLMET  & \begin{tabular}[c]{@{}l@{}}person (.992), time (.314), title (.100)...\\ ...consortium (.000), negotiator (.000), football player (.000)\end{tabular} \\\bottomrule
\end{tabular}%
}
\caption{Top and bottom predictions and their probabilities when querying typing models with empty input.}
\label{tab:overgeneralization}
\end{table}

\begin{table*}[t!]
\setlength{\tabcolsep}{4pt}
\resizebox{\textwidth}{!}{%
\begin{tabular}{@{}lcccccc|cccccc@{}}\toprule
\multirow{2}{*}{Approach} & \multicolumn{6}{c}{BiLSTM}                                                                                                                                & \multicolumn{6}{c}{MLMET}                                                                                                      \\\cmidrule(r){2-13}
                          & U-Prec.             & U-Rec.                & U-F1                    & A-Prec.             & A-Rec.                & A-F1                    & U-Prec.    & U-Rec.                & U-F1           & A-Prec.    & A-Rec.                & A-F1                    \\\midrule
No Debiasing              & .471                   & .242                   & .320                   & .242                   & .182                   & .208                   & \textbf{.527} & .452                   & .487          & \textbf{.554} & .414                   & .474                   \\\midrule
AFLite                    & \textit{\textbf{.529}} & .209                   & .300                   & \textit{.296}          & .142                   & .192                   & .526          & .450                   & .485          & .551          & \textit{.420}          & \textit{.477}          \\
POE                       & .441                   & \textit{.281}          & \textit{\textbf{.343}} & \textit{.339}          & \textit{.267}          & \textit{.299}          & .518          & \textit{.458}          & .486          & .543          & \textit{.425}          & \textit{.477}          \\
Focal                     & .315                   & \textit{\textbf{.341}} & \textit{.328}          & .207                   & \textit{.269}          & \textit{.234}          & .520          & \textit{.460}          & \textit{.488} & .545          & \textit{.419}          & .474                   \\
Learned-mixin             & .396                   & \textit{.289}          & \textit{.335}          & \textit{.282}          & \textit{.279}          & \textit{.280}          & .483          & \textit{.472}          & .478          & .480          & \textit{.453}          & .466                   \\
Learned-mixin+H           & .279                   & \textit{.326}          & .301                   & .182                   & \textit{.338}          & \textit{.237}          & .405          & \textit{\textbf{.529}} & .459          & .379          & \textit{.485}          & .425                   \\\midrule
Contrastive (CE)          & .461                   & \textit{.272}          & \textit{.342}          & \textit{.312}          & \textit{.354}          & \textit{.331}          & .477          & \textit{.471}          & .474          & .449          & \textit{.460}          & .454                   \\
Contrastive (Cosine)      & .440                   & \textit{.257}          & \textit{.325}          & \textit{\textbf{.462}} & \textit{.265}          & \textit{.337}          & .495          & .451                   & .472          & .489          & \textit{.441}          & .464                   \\\midrule
Counterfact. Inf.            & .442                   & \textit{.264}          & \textit{.331}          & \textit{.347}          & .173                   & \textit{.231}          & .525          & \textit{.454}          & .487          & .492          & \textit{.446}          & .468                   \\\midrule\midrule
Augmentation (ours)       & \textit{.473}          & \textit{.260}          & \textit{.336}          & \textit{.345}          & \textit{\textbf{.367}} & \textit{\textbf{.356}} & .515          & \textit{.466}          & \textbf{.489} & .540          & \textit{\textbf{.470}} & \textit{\textbf{.530}}\\\bottomrule
\end{tabular}%
}
\vspace{-0.5em}
\caption{Effectiveness of different debiasing approaches on two representative entity typing models when testing on UFET test set (U-) and our counterfactual augmented test set (A-). 
The best performance per column is marked in \textbf{bold} while improved values over those without debiasing in \textit{italic}. For contrastive learning, CE stands for the cross entropy and Cosine represents cosine similarity. See results of three other entity typing models in~\Cref{tab:mitigation_cont}.}
\label{tab:mitigation}
\vspace{-0.5em}
\end{table*}

\subsection{Mitigating Spurious Correlations}\label{sec:mitigating}
In~\Cref{tab:mitigation}, we evaluate robustness of entity typing models after adopting the proposed counterfactual data augmentation or alternative debiasing techniques, and present results on the \emph{UFET} test set with bias and our counterfactually 
debiased test set.

Overall, our counterfactual data augmentation is the only approach that consistently improves the generalization of the studied models across both test sets. Particularly, we achieve the best performance on \emph{UFET} and the debiased test set with \emph{MLMET}. Besides, models trained with our approach improve the performance of \emph{BiLSTM} and \emph{MLMET} relatively by $71.15\%$ and $11.81\%$ on the debiased test set, respectively, implying the least reliance on spurious correlations to infer correct entity types.

When evaluating other debiasing approaches, we find that 1) none of the 
resampling or reweighting techniques 
is capable to maintain the performance on UFET test set of both models, which could be attributed to the large-scale label space and the existence of diverse causes of model biases; 2) contrastive learning with either cross entropy loss or cosine similarity loss helps improve performance on debiased samples, but leads to accuracy drop of \emph{MLMET} on UFET; 3) without updating model parameters given bias features, counterfactual inference fails to improve performance  of \emph{MLMET} on debiased samples. 

\section{Related Work}


\stitle{Entity Typing}
Earlier studies on entity typing \cite{yogatama-etal-2015-embedding,ren2016label,xu-barbosa-2018-neural} learned contextual embeddings for entity mentions 
and types to capture their association.
To model label correlations without annotated label hierarchies in UFET, LabelGCN~\cite{xiong-etal-2019-imposing} introduced the graph propagation layer to encode global label co-occurrence statistics and their word-level similarities,
whereas HMGCN~\cite{jin-etal-2019-fine} proposed to infer this information from a knowledge base. 
For the same purpose,
Box4Types~\cite{onoe-etal-2021-modeling} was proposed to embed concepts as hyper rectangles (boxes), 
such that their topological relations can represent type hierarchies.
Considering the prevailing noisy labels in existing entity typing datasets,
\citet{onoe-durrett-2019-learning} performed supervised denoising 
to filter and fix noisy training labels.
\citet{dai-etal-2019-improving} 
introduced distant supervision from entity linking results.
To tackle the sparsity of training, recent work conducted data augmentation with a masked language model and WordNet knowledge to enrich the training data~(\citealt{dai-etal-2021-ultra}; MLMET), 
and made use indirect supervision from natural language inference~(\citealt{li-etal-2022-ultra}; LITE).
Despite much attention in literature, to the best of our knowledge, our work represents the first investigation on faithfulness and reducing shortcuts in this task.

\stitle{Spurious Correlations in NLP Models}
Much recent effort has been put into studying spurious correlation in Natural Language Inference (NLI) tasks.
Recent studies show that crowd workers 
are prone to produce annotation artifacts ~\cite{gururangan-etal-2018-annotation} through the rapid annotation process and result in identifiable shortcut features \cite{karimi-mahabadi-etal-2020-end,du-etal-2021-towards}.
Hence, simple models can easily achieve good performance
even with partial inputs
\cite{kaushik2019learning,karimi-mahabadi-etal-2020-end}, 
or leveraging superficial syntactic properties 
\cite{mccoy-etal-2019-right,utama-etal-2020-mind,pezeshkpour2021combining}. 
On several other NLP tasks composed of multiple textual components, it has been observed that models fed with partial inputs can already achieve competitive performance, 
e.g., 
predicting for claim verification \cite{schuster-etal-2019-towards,utama-etal-2020-mind,du2021towards} or argument reasoning comprehension~\cite{niven-kao-2019-probing,branco-etal-2021-shortcutted} with only the claim, choosing a plausible story ending 
without seeing the story~\cite{cai-etal-2017-pay}, 
question answering 
using a positional bias~\cite{jia-liang-2017-adversarial,kaushik-lipton-2018-much}, etc.

The spurious correlation problems in information extraction tasks are still an under-explored area.
Despite most recent studies on NER \cite{zhang-etal-2021-de} and relation extraction \cite{wang2022should},
this work represents the first attempt to diagnose spurious correlations in entity typing, for which we comprehensively analyzed various types of causes for biases and provided a dedicated debiasing method.
We also conducted a comprehensive comparison with various alternatives based on resampling \cite{le2020adversarial}, reweighting \cite{clark-etal-2019-dont,karimi-mahabadi-etal-2020-end} and counterfactual inference \cite{wang2022should}. 

\section{Conclusions}
To comprehensively investigate the faithfulness and reliability of entity typing methods, we systematically define six kinds of model biases that are reflected mainly from spurious correlations. 
In addition to diagnosing the biases on representative models using benchmark data, we also present
a counterfactual data augmentation approach that helps improve the generalization of different entity typing models with consistently better performance on both original and debiased test sets.
\section*{Limitations}
There are two important caveats to this work. First, for instances identified with a particular bias by the PLM, we do not guarantee all typing models would exploit spurious correlations on it. To the best of our knowledge, entity typing models with spurious correlation ablated and mitigated do not yet exist. Although we observe significant performance differences between the original biased instances and the crafted debiased counterparts from existing entity typing models, we hope future work would pay attention to spurious correlations, and develop models with improved robustness and generalization performance. Second, although biases defined in this work  comprehensively cover six aspects, but still they may not exhaust 
all kinds of biased prediction in entity typing. 
In our study we only tried our best effort to study the most noteworthy and typical biases with which models may inflate performance by leveraging corresponding spurious correlations. At the same time, appeal for more research efforts to complete our understanding with more biases investigated.
In addition, the studied model biases are representative to the widely practiced classification-based typing paradigm. There are effects in the most recent NLI-based or bi-encoder-based methods \cite{li-etal-2022-ultra,huang-etal-2022-unified}, which require further analysis.

\section*{Ethical Consideration}
We acknowledge the importance of ethical considerations in language technologies and would like to point the reader to the following concern. Gender is a spectrum and we respect all gender identities, e.g., nonbinary, genderfluid, polygender, omnigender, etc. To craft instances free from pronoun bias, we substitute the gender pronouns with their referred names in contexts if they exist, or random masculine/feminine given names otherwise. This is due to the lack of entity typing datasets going beyond binarism for pronoun mentions such as they/them/theirs, ze/hir/hir, etc. Nevertheless, we support the rise of alternative neutral pronoun expressions and look forward to the development of non-binary inclusive datasets and technologies.
In the meantime, although our techniques do not introduce or exaggerate possible gender bias in the original experimental data, in cases where such biases pre-exist in those data, additional gender neuralization techniques would be needed in order for such biases to be mitigated.

\section*{Acknowledgement}
We appreciate the anonymous reviewers for their insightful comments and suggestions. This material
is partly supported by the National Science Foundation
of United States Grant IIS 2105329 and a Cisco Research Award. 
Nan Xu and Fei Wang are supported by the Annenberg Fellowships.
Bangzheng Li is supported by the USC Provost's Ph.D. Fellowship.
Mingtao Dong is supported by the USC Provost's Undergrad Research Fellowship.

\bibliography{anthology,custom}
\bibliographystyle{acl_natbib}
\clearpage
\appendix
\section{Appendix}\label{sec:appendix}

\subsection{Additional Details about Mitigating Spurious Correlations}~\label{sec:mitigate_examples}
\paragraph{Lexical Overlapping Bias} 
We consider the following sentence as an instance: ``Deutsche Bank would neither confirm nor deny the \underline{\textbf{\emph{discharge}}} of the two executives, and it also would not specify who was the target of the alleged spying'', annotated with types \emph{dismissal, discharge, leave, termination}. Since ``discharge'' shows up both in the mention and the true labels, we perform word substitutions with synonym candidates from 20 synsets found in WordNet. We show a few synsets with popular senses as follows:

\quad$\mathbb{\bf Synset}$ I: (\emph{the termination of someone's employment}) dismissal, dismission, discharge, firing, liberation, release, sack, sacking

\quad$\mathbb{\bf Synset II}$: (\emph{a substance that is emitted or released}) discharge, emission

\quad$\mathbb{\bf Synset III}$: (\emph{a formal written statement of relinquishment}) release, waiver, discharge

Synonyms that share high word similarities with the true labels are removed to avoid creating new lexical overlapping bias features, e.g., \emph{dismissal, discharge} from Synset I, \emph{discharge} from Synset II and Synset III. To guarantee the semantic consistency of the new sentence and the fidelity of true labels to type the new mention, we leverage available word sense disambiguation models to preserve synonyms from the synset that is most consistent with the sense used in the original sentence: \emph{dismission, firing, liberation, release, sack, sacking} from Synset I are finally selected to substitute ``discharge''. As shown in \hyperlink{t4}{T4} of~\Cref{tab:bias_example_I}, without training on the debiased set, \emph{MLMET} no longer predicts the overlapped type ``day'', but some surface word ``period'' instead.
\paragraph{Named Entity Bias}
Compared with the politician \emph{Benjamin Netanyahu}, the PLM can hardly infer the impression of the singer \emph{Jintara Poonlarp} on the public. Particularly, only general types to describe person named entities are predicted in \hyperlink{s5}{S5}: \emph{person, human, woman}. We then consider \emph{Benjamin Netanyahu} as a biased named entity containing much prior knowledge, while \emph{Jintara Poonlarp} as an unbiased named entity without much type-relevant information revealed. After substituting \emph{Benjamin Netanyahu} with \emph{Jintara Poonlarp} in \hyperlink{t6}{T6}, \emph{MLMET} can hardly infer the political role of the new mention by analyzing its connection with the politician (\emph{Amin al-Husseini}, Palestinian Arab nationalist and Muslim leader in Mandatory Palestine\footnote{\url{https://en.wikipedia.org/wiki/Amin\_al-Husseini}}) and political description (``masterminds'' and ``Holocaust'') in the context. \emph{MLMET} even crashes with some out-of-context predictions: \emph{scholar, writer}.
\paragraph{Pronoun Bias}
As shown in the original instance \hyperlink{t7}{T7} of~\Cref{tab:bias_example_I}, the actual person's name that the pronoun mention ``Her'' refers to is not  provided in the current sentence. As a result, a random feminine name, ``Judith'' is assumed to be the referred entity and substitutes the pronoun mention as a new sentence in \hyperlink{s6}{S6}. Considering the ridiculously wrong types predicted by \emph{RoBERTa} such as \emph{bird} and \emph{cat}, we include this new instance in the debiased set and expect the entity modeling training on this kind of instances to infer person name types as accurate as pronoun types. Beforehand, we test on the newly crafted instance without counterfactual augmented training, and observe huge performance drop after pronoun concretization: types related to the name's gender attribute such as \emph{woman} and \emph{female} are missing, let alone the types requiring fully context understanding such as \emph{actress}. 
\paragraph{Dependency Bias}
For instance \hyperlink{t9}{T9} in~\Cref{tab:bias_example_II}, we show their mention word dependency analysis in \hyperlink{s6}{S6} and predictions on the perturbed instance in \hyperlink{t10}{T10}. Without distractions from other dependent words in the new mention, \emph{MLMET} spares no effort to infer types of the target entity ``whale'' with the correct prediction \emph{subject}. Motivated by the improved performance when the mention headword is specifically provided, we believe entity typing models can actively learn to capture target entity among mention words when both original sentences and their debiased counterparts are given during training. In such augmented training regime, the entity typing model is expected achieve robust performance on new sentences bearing distractions from dependent words in mentions.

\subsection{Implementation Details}
We adopt the released checkpoints of RoBERTa-large~\cite{liu2019roberta} as the PLM to identify biased instances. To perform masked fill-in, we adopt the top 10 predictions and filter out non-type words as the predicted types. We recognize potentially biased samples based on PLM predictions based on the following criteria. \emph{1) mention-context bias}: instances are considered biased if the PLM can predict the type labels with the F1 score above $0.5$ when only the mention is provided; \emph{2) named entity bias}: instances are considered biased if the PLM can predict types labels with the F1 score above $0.5$ when only the named entity is given; \emph{3) lexical overlapping bias}: instances are considered biased if the PLM makes predictions with the F1 score below $0.5$ after substituting overlapped words with their semantically similar words; \emph{4) pronoun bias}: for pronouns without coreferenced entities detected, we substitute them with 5 random real person names as debiased instances. Instances are considered biased if the PLM makes predictions with the F1 score below $0.5$ after real name substitution. We mainly use $0.5$ as the threshold to distinguish biased samples from unbiased, since the SOTA model achieves the F1 score approximating $0.5$ on average of the UFET test samples.

To diagnose entity typing models, for those with released checkpoints (\emph{BiLSTM}, \emph{Box4Types}, \emph{LRN}), we directly evaluate on the original (un)biased and crafted debiased instances. We train \emph{LabelGCN} and \emph{MLMET} by ourselves following hyperparameters and training strategies introduced in their papers. 

To evaluate various debiasing approaches, we train entity typing models using checkpoints training on the original dataset as the warm start with the same hyperparameter sets. 

We run experiments on a commodity server with a GeForce RTX 2080 GPU. It takes about 4 hours to train one entity typing model on average and 2 minutes for inference on the UFET test set.
\subsection{OntoNotes Experiments}~\label{sec:ontonotes}
We diagnose entity typing models and the effectiveness of the proposed counterfactual augmented approach on OntoNotes~\cite{gillick2014context}. The original dataset contains $251,309$ instances automatically annotated by linking identified entity mentions to Freebase profiles for training, and $11,165$ manually annotated instances: $2,202$ for validation and $8,963$ for testing, respectively. Its label space is constituted of  $89$ types organized into a hierarchy, e.g., /person (level 1), /person/artist (level 2), /person/artist/actor (level 3). We adopt the set augmented by~\cite{choi-etal-2018-ultra} for model training: $793,487$ instances with distant supervision from Wikipedia definition sentences and head word supervision.

In~\Cref{tab:diagnosing_ontonotes}, we report performance of two representative entity typing models on original biased samples where they are likely to exploit spurious correlations, the perturbed counterparts, as well as performance on unbiased samples. We have the following observations: 1) entity typing models can achieve satisfactory performance when only the mention is provided without context; 2) considering lexical overlapping bias, performance on both biased and unbiased samples identified by the PLM drops a lot after substituting overlapped mention words with their sematically similar words; 3) the performance variation after named entity substitution is evident; 4) models can obtain much better performance on some instances when the headwords are explicitly given without distractions from other words in mentions; 5) performance on instances purely annotated by coarse and fine labels is good in general with around $15\%$ difference in F1 score. Similarly to UFET, models training on OntoNotes may achieve good performance without reasoning on the context, rely on lexical overlapping between mention words and types to make precise predictions, and obtain below-average results on some instances for lack of syntactic structure understanding.

To mitigate spurious correlations, we evaluate the proposed counterfactual augmented approach in~\Cref{tab:mitigation_ontonotes}. With additional debiased instances for model training, both \emph{BiLSTM} and \emph{MLMET} maintain good performance on the original OntoNotes test set and much higher accuracy on the corresponding debiased test set, leading to improved generalization.

\begin{table*}[t!]
\resizebox{\linewidth}{!}{%
\setlength{\tabcolsep}{3pt}
\begin{tabular}{@{}lccccccccccccccc@{}}
\toprule
\multirow{2}{*}{Test Set} & \multicolumn{3}{c}{BiLSTM}                                                                       & \multicolumn{3}{c}{MLMET}                                   & \multicolumn{3}{c}{LabelGCN}                                                         & \multicolumn{3}{c}{Box4Types}& \multicolumn{3}{c}{LRN}                                                                                                                                                     \\ \cmidrule(l){2-16} 
                          & Prec.                                               & Rec.                                                 & F1          & Prec.                                               & Rec.                                                 & F1 & Prec.                                               & Rec.                                                 & F1 & Prec.                                               & Rec.                                                 & F1                                             & Prec.                                               & Rec.                                                  & F1                                                      \\ \midrule
 \multicolumn{16}{c}{\textbf{Mention-Context Bias} ($\uparrow$)} \\\cmidrule(l){7-10} 
Biased                    & .602                                                    & .283                                                    & .385                                                    & .668                                                    & .640                                                     & .654   &.689&.418&.521&.679&.516&586 &.666&.425&.519                                                \\
-Perturb.                 & \begin{tabular}[c]{@{}c@{}}.606\\ (0.7\%)\end{tabular}  & \begin{tabular}[c]{@{}c@{}}.298\\ (5.3\%)\end{tabular}  & \begin{tabular}[c]{@{}c@{}}.400\\ (3.8\%)\end{tabular}  & \begin{tabular}[c]{@{}c@{}}.682\\ (2.1\%)\end{tabular}  & \begin{tabular}[c]{@{}c@{}}.628\\ (-1.9\%)\end{tabular}  & \begin{tabular}[c]{@{}c@{}}.654\\ (0.0\%)\end{tabular} &\begin{tabular}[c]{@{}c@{}}.699\\ (1.3\%)\end{tabular}&\begin{tabular}[c]{@{}c@{}}.398\\ (-4.9\%)\end{tabular}&\begin{tabular}[c]{@{}c@{}}.507\\ (-2.6\%)\end{tabular}&\begin{tabular}[c]{@{}c@{}}.644\\ (-5.2\%)\end{tabular}&\begin{tabular}[c]{@{}c@{}}.486\\ (-5.8\%)\end{tabular}&\begin{tabular}[c]{@{}c@{}}.554\\ (-5.5\%)\end{tabular}&\begin{tabular}[c]{@{}c@{}}.653\\ (-2.0\%)\end{tabular}&\begin{tabular}[c]{@{}c@{}}.434\\ (2.2\%)\end{tabular}&\begin{tabular}[c]{@{}c@{}}.522\\ (0.5\%)\end{tabular} \\\midrule
Unbiased                  & .452                                                    & .188                                                    & .265                                                    & .486                                                    & .395                                                     & .436 &.484&.235&.317&.470&.319&.380&.563&.272&.367                                              \\
-Perturb.                 & \begin{tabular}[c]{@{}c@{}}.450\\ (-0.5\%)\end{tabular} & \begin{tabular}[c]{@{}c@{}}.176\\ (-6.4\%)\end{tabular} & \begin{tabular}[c]{@{}c@{}}.253\\ (-4.7\%)\end{tabular} & \begin{tabular}[c]{@{}c@{}}.453\\ (-6.8\%)\end{tabular} & \begin{tabular}[c]{@{}c@{}}.350\\ (-11.4\%)\end{tabular} & \begin{tabular}[c]{@{}c@{}}.395\\ (-9.4\%)\end{tabular}&\begin{tabular}[c]{@{}c@{}}.459\\ (-5.2\%)\end{tabular} &\begin{tabular}[c]{@{}c@{}}.221\\ (-6.1\%)\end{tabular}&\begin{tabular}[c]{@{}c@{}}.298\\ (-5.8\%)\end{tabular}&\begin{tabular}[c]{@{}c@{}}.446\\ (-5.1\%)\end{tabular}&\begin{tabular}[c]{@{}c@{}}.279\\ (-12.6\%)\end{tabular}&\begin{tabular}[c]{@{}c@{}}.343\\ (-9.8\%)\end{tabular}&\begin{tabular}[c]{@{}c@{}}.509\\ (-9.6\%)\end{tabular}&\begin{tabular}[c]{@{}c@{}}.232\\ (-14.7\%)\end{tabular}&\begin{tabular}[c]{@{}c@{}}.319\\ (-13.1\%)\end{tabular} \\\midrule
 \multicolumn{16}{c}{\textbf{Lexical Overlapping Bias} ($\downarrow$)} \\\cmidrule(l){7-10} 
Biased                                                                            & .415                                                     & .661                                                     & .510                                                     & .641                                                     & .484                                                     & .551  &.651&.292&.403 &.118&.442&.186 &.089&.259&.133                                                 \\
-Perturb.                                                                         & \begin{tabular}[c]{@{}c@{}}.040\\ (-90.4\%)\end{tabular} & \begin{tabular}[c]{@{}c@{}}.065\\ (-90.1\%)\end{tabular} & \begin{tabular}[c]{@{}c@{}}.050\\ (-90.3\%)\end{tabular} & \begin{tabular}[c]{@{}c@{}}.411\\ (-35.8\%)\end{tabular} & \begin{tabular}[c]{@{}c@{}}.271\\ (-44.0\%)\end{tabular} & \begin{tabular}[c]{@{}c@{}}.327\\ (-40.8\%)\end{tabular}&\begin{tabular}[c]{@{}c@{}}.338\\ (-48.1\%)\end{tabular}&\begin{tabular}[c]{@{}c@{}}.106\\ (-63.6\%)\end{tabular}&\begin{tabular}[c]{@{}c@{}}.162\\ (-59.9\%)\end{tabular}&\begin{tabular}[c]{@{}c@{}}.074\\ (-36.9\%)\end{tabular}&\begin{tabular}[c]{@{}c@{}}.251\\ (-43.3\%)\end{tabular}&\begin{tabular}[c]{@{}c@{}}.115\\ (-38.3\%)\end{tabular}&\begin{tabular}[c]{@{}c@{}}.066\\ (-25.8\%)\end{tabular}&\begin{tabular}[c]{@{}c@{}}.158\\ (-39.0\%)\end{tabular}&\begin{tabular}[c]{@{}c@{}}.093\\ (-29.7\%)\end{tabular} \\\midrule
Unbiased                                                                          & .278                                                     & .667                                                     & .392                                                     & .522                                                     & .567                                                     & .544&.667&.300&.414&.167&.667&.267&.067&.333&.111                                                     \\
-Perturb.                                                                         & \begin{tabular}[c]{@{}c@{}}.111\\ (-60.0\%)\end{tabular} & \begin{tabular}[c]{@{}c@{}}.333\\ (-50.0\%)\end{tabular} & \begin{tabular}[c]{@{}c@{}}.167\\ (-57.5\%)\end{tabular} & \begin{tabular}[c]{@{}c@{}}.381\\ (-27.1\%)\end{tabular} & \begin{tabular}[c]{@{}c@{}}.533\\ (-5.9\%)\end{tabular}  & \begin{tabular}[c]{@{}c@{}}.444\\ (-18.2\%)\end{tabular}&\begin{tabular}[c]{@{}c@{}}.333\\ (-50.0\%)\end{tabular}&\begin{tabular}[c]{@{}c@{}}.067\\ (-77.8\%)\end{tabular}&\begin{tabular}[c]{@{}c@{}}.111\\ (-73.1\%)\end{tabular}&\begin{tabular}[c]{@{}c@{}}.222\\ (33.3\%)\end{tabular}&\begin{tabular}[c]{@{}c@{}}.667\\ (0.0\%)\end{tabular}&\begin{tabular}[c]{@{}c@{}}.333\\ (25.0\%)\end{tabular}&\begin{tabular}[c]{@{}c@{}}.056\\ (-16.7\%)\end{tabular}&\begin{tabular}[c]{@{}c@{}}.333\\ (0.0\%)\end{tabular}&\begin{tabular}[c]{@{}c@{}}.095\\ (-14.3\%)\end{tabular} \\ \midrule
 \multicolumn{16}{c}{\textbf{Named Entity Bias} ($\downarrow$)} \\\cmidrule(l){7-10} 
Biased                                                                  & .744                                                     & .380                                                     & .504                                                     & .719                                                     & .752                                                     & .735&.730&.524&.610&.686&.658&.671&.754&.557&.641                                                     \\
-Perturb.                                                               & \begin{tabular}[c]{@{}c@{}}.538\\ (-27.7\%)\end{tabular} & \begin{tabular}[c]{@{}c@{}}.240\\ (-36.9\%)\end{tabular} & \begin{tabular}[c]{@{}c@{}}.332\\ (-34.0\%)\end{tabular} & \begin{tabular}[c]{@{}c@{}}.615\\ (-14.5\%)\end{tabular} & \begin{tabular}[c]{@{}c@{}}.586\\ (-22.0\%)\end{tabular} & \begin{tabular}[c]{@{}c@{}}.600\\ (-18.4\%)\end{tabular}&\begin{tabular}[c]{@{}c@{}}.568\\ (-22.2\%)\end{tabular}&\begin{tabular}[c]{@{}c@{}}.362\\ (-30.9\%)\end{tabular}&\begin{tabular}[c]{@{}c@{}}.442\\ (-27.5\%)\end{tabular}&\begin{tabular}[c]{@{}c@{}}.541\\ (-21.2\%)\end{tabular}&\begin{tabular}[c]{@{}c@{}}.479\\ (-27.2\%)\end{tabular}&\begin{tabular}[c]{@{}c@{}}.508\\ (-24.4\%)\end{tabular}&\begin{tabular}[c]{@{}c@{}}.634\\ (-15.9\%)\end{tabular}&\begin{tabular}[c]{@{}c@{}}.448\\ (-19.5\%)\end{tabular}&\begin{tabular}[c]{@{}c@{}}.525\\ (-18.0\%)\end{tabular} \\\midrule
Unbiased                                                                & .522                                                     & .226                                                     & .316                                                     & .536                                                     & .477                                                     & .505&.542&.288&.376&.520&.407&.457&.595&.365&.453                                                    \\
-Perturb.                                                               & \begin{tabular}[c]{@{}c@{}}.613\\ (17.5\%)\end{tabular}  & \begin{tabular}[c]{@{}c@{}}.267\\ (17.7\%)\end{tabular}  & \begin{tabular}[c]{@{}c@{}}.372\\ (17.6\%)\end{tabular}  & \begin{tabular}[c]{@{}c@{}}.582\\ (8.6\%)\end{tabular}   & \begin{tabular}[c]{@{}c@{}}.502\\ (5.4\%)\end{tabular}   & \begin{tabular}[c]{@{}c@{}}.539\\ (6.9\%)\end{tabular}&\begin{tabular}[c]{@{}c@{}}.651\\ (20.2\%)\end{tabular}&\begin{tabular}[c]{@{}c@{}}.337\\ (17.3\%)\end{tabular}&\begin{tabular}[c]{@{}c@{}}.444\\ (18.2\%)\end{tabular}&\begin{tabular}[c]{@{}c@{}}.538\\ (3.4\%)\end{tabular}&\begin{tabular}[c]{@{}c@{}}.472\\ (16.1\%)\end{tabular}&\begin{tabular}[c]{@{}c@{}}.503\\ (10.2\%)\end{tabular}&\begin{tabular}[c]{@{}c@{}}.629\\ (5.8\%)\end{tabular}&\begin{tabular}[c]{@{}c@{}}.433\\ (18.6\%)\end{tabular}&\begin{tabular}[c]{@{}c@{}}.513\\ (13.4\%)\end{tabular}  \\ \midrule
 \multicolumn{16}{c}{\textbf{Pronoun Bias} ($\downarrow$)} \\\cmidrule(l){7-10} 

Biased                   & .567                                                     & .438                                                     & .494                                                     & .555                                                    & .566                                                     & .561&.555&.474&.511&.576&.555&.565&.716&.474&.570                                               \\
-Perturb.                & \begin{tabular}[c]{@{}c@{}}.148\\ (-74.0\%)\end{tabular} & \begin{tabular}[c]{@{}c@{}}.227\\ (-48.2\%)\end{tabular} & \begin{tabular}[c]{@{}c@{}}.179\\ (-63.8\%)\end{tabular} & \begin{tabular}[c]{@{}c@{}}.619\\ (11.5\%)\end{tabular} & \begin{tabular}[c]{@{}c@{}}.455\\ (-19.6\%)\end{tabular} & \begin{tabular}[c]{@{}c@{}}.525\\ (-6.4\%)\end{tabular}&\begin{tabular}[c]{@{}c@{}}.578\\ (4.2\%)\end{tabular}&\begin{tabular}[c]{@{}c@{}}.300\\ (-36.7\%)\end{tabular}&\begin{tabular}[c]{@{}c@{}}.395\\ (-22.8\%)\end{tabular}&\begin{tabular}[c]{@{}c@{}}.481\\ (-16.5\%)\end{tabular}&\begin{tabular}[c]{@{}c@{}}.397\\ (-28.4\%)\end{tabular}&\begin{tabular}[c]{@{}c@{}}.435\\ (-23.0\%)\end{tabular}&\begin{tabular}[c]{@{}c@{}}.776\\ (8.2\%)\end{tabular}&\begin{tabular}[c]{@{}c@{}}.415\\ (-12.3\%)\end{tabular}&\begin{tabular}[c]{@{}c@{}}.541\\ (-5.2\%)\end{tabular} \\\midrule
Unbiased                 & .405                                                     & .334                                                     & .366                                                     & .738                                                    & .635                                                     & .683&.660&.510&.575&.715&.759&.736&.633&.346&.448                                                    \\
-Perturb.                & \begin{tabular}[c]{@{}c@{}}.106\\ (-73.9\%)\end{tabular} & \begin{tabular}[c]{@{}c@{}}.167\\ (-50.0\%)\end{tabular} & \begin{tabular}[c]{@{}c@{}}.130\\ (-64.6\%)\end{tabular} & \begin{tabular}[c]{@{}c@{}}.735\\ (-0.5\%)\end{tabular} & \begin{tabular}[c]{@{}c@{}}.597\\ (-6.0\%)\end{tabular}  & \begin{tabular}[c]{@{}c@{}}.659\\ (-3.5\%)\end{tabular}&\begin{tabular}[c]{@{}c@{}}.703\\ (6.6\%)\end{tabular}&\begin{tabular}[c]{@{}c@{}}.258\\ (-49.4\%)\end{tabular}&\begin{tabular}[c]{@{}c@{}}.378\\ (-34.4\%)\end{tabular}&\begin{tabular}[c]{@{}c@{}}.670\\ (-6.3\%)\end{tabular}&\begin{tabular}[c]{@{}c@{}}.548\\ (-27.8\%)\end{tabular}&\begin{tabular}[c]{@{}c@{}}.603\\ (-18.1\%)\end{tabular}&\begin{tabular}[c]{@{}c@{}}.555\\ (-12.4\%)\end{tabular}&\begin{tabular}[c]{@{}c@{}}.339\\ (-2.1\%)\end{tabular}&\begin{tabular}[c]{@{}c@{}}.421\\ (-6.0\%)\end{tabular} \\ \midrule
 \multicolumn{16}{c}{\textbf{Dependency Bias}} \\\cmidrule(l){7-10}
UFET                                                                & .350                                                    & .097                                                     & .152                                                     & .450                                                    & .402                                                    & .424&.452&.248&.321&.462&.351&.399&.491&.300&.372                                                    \\
-Perturb.                                                                 & \begin{tabular}[c]{@{}c@{}}.407\\ (16.2\%)\end{tabular} & \begin{tabular}[c]{@{}c@{}}.386\\ (297.3\%)\end{tabular} & \begin{tabular}[c]{@{}c@{}}.396\\ (160.5\%)\end{tabular} & \begin{tabular}[c]{@{}c@{}}.617\\ (37.1\%)\end{tabular} & \begin{tabular}[c]{@{}c@{}}.520\\ (29.4\%)\end{tabular} & \begin{tabular}[c]{@{}c@{}}.564\\ (32.9\%)\end{tabular}&\begin{tabular}[c]{@{}c@{}}.757\\ (67.3\%)\end{tabular}&\begin{tabular}[c]{@{}c@{}}.392\\ (58.1\%)\end{tabular}&\begin{tabular}[c]{@{}c@{}}.517\\ (61.2\%)\end{tabular}&\begin{tabular}[c]{@{}c@{}}.710\\ (53.7\%)\end{tabular}&\begin{tabular}[c]{@{}c@{}}.482\\ (37.2\%)\end{tabular}&\begin{tabular}[c]{@{}c@{}}.574\\ (43.9\%)\end{tabular}&\begin{tabular}[c]{@{}c@{}}.715\\ (45.7\%)\end{tabular}&\begin{tabular}[c]{@{}c@{}}.445\\ (48.5\%)\end{tabular}&\begin{tabular}[c]{@{}c@{}}.549\\ (47.4\%)\end{tabular} \\\midrule
 \multicolumn{16}{c}{\textbf{Overgeneralization Bias}} \\\cmidrule(l){7-10}
Coarse     & .362                                                    & .656                                                     & .466                                                     & .297                                                    & .758                                                    & .427&.360&.656&.465&.339&.688&.454&.466&.683&.554                                                    \\
Ultra-fine & .157                                                    & .094                                                     & .118                                                     & .221                                                    & .244                                                    & .232 &.207&.120&.152&.147&.120&.132&.161&.091&.116                                                   \\ \bottomrule
\end{tabular}%
}
\caption{Performance of all entity typing models evaluated by complete metrics (Prec. for precision, Rec. for recall and F1 for F1 score) on UFET testing samples with(out) distinct bias and their perturbations.}
\label{tab:ufet_full}
\end{table*}

\begin{table*}[h]
 \begin{minipage}{\textwidth}

\resizebox{\textwidth}{!}{%
\setlength{\tabcolsep}{3pt}
\begin{tabular}{@{}lcccccc|cccc@{}}
\toprule
\multirow{2}{*}{Test set} & \multicolumn{2}{c}{\begin{tabular}[c]{@{}c@{}}Mention-Context \\ 1830/7133 ($\uparrow$)\end{tabular}}                                                                                   & \multicolumn{2}{c}{\begin{tabular}[c]{@{}c@{}}Lexical Overlapping\\234/319 ($\downarrow$) \end{tabular}}                                      & \multicolumn{2}{c|}{\begin{tabular}[c]{@{}c@{}}Named Entity\\1132/2457 ($\downarrow$) \end{tabular}}                                                                                                                                                                                    & \multicolumn{2}{c}{\begin{tabular}[c]{@{}c@{}}Dependency\\842,544/7129  \end{tabular}}                                                                                        & \multicolumn{2}{c}{\begin{tabular}[c]{@{}c@{}}Overgeneralization\\5549/3414  \end{tabular}}                                                                                  \\
\cmidrule(r){2-11}
                          & BiLSTM                                                    & MLMET                                                     & BiLSTM                                                     & MLMET                                                      & BiLSTM                                                     & MLMET                                                      & BiLSTM                                                     & MLMET                                                     & BiLSTM                                                    & MLMET                                                    \\ \midrule
Biased                &  .719                                                   & .821                                                   &  .922                                                   & .940                                                     & .774                                                     & .844                                                                                                                                                                &  .291                                                   &  .416                                                  &   .844                                                   & .909                                                      \\
-Perturb.             & \begin{tabular}[c]{@{}c@{}} .698\\ (-3.0\%)\end{tabular}  & \begin{tabular}[c]{@{}c@{}} .803\\ (-2.2\%)\end{tabular}  & \begin{tabular}[c]{@{}c@{}}.345\\ (-62.6\%)\end{tabular}  & \begin{tabular}[c]{@{}c@{}} .473\\ (-49.7\%)\end{tabular} & \begin{tabular}[c]{@{}c@{}} .668\\ (-13.7\%)\end{tabular} & \begin{tabular}[c]{@{}c@{}} .765\\ (-9.4\%)\end{tabular} &   \begin{tabular}[c]{@{}c@{}}.808\\ (178.0\%)\end{tabular} & \begin{tabular}[c]{@{}c@{}} .909\\ (118.5\%)\end{tabular} &  .646                                                         & .757                                                          \\\midrule
Debiased                &  .787                                                   &  .864                                                   &  .982                                                    & .983                                                      & .761                                                     &.855                                                                                                           &  -                                                         & -                                                         & -&- \\
-Perturb.             & \begin{tabular}[c]{@{}c@{}} .787\\ (.1\%)\end{tabular} & \begin{tabular}[c]{@{}c@{}} .847\\ (-2.0\%)\end{tabular} & \begin{tabular}[c]{@{}c@{}} .407\\ (-58.6\%)\end{tabular} & \begin{tabular}[c]{@{}c@{}} .467\\ (-52.5\%)\end{tabular} & \begin{tabular}[c]{@{}c@{}} .665\\ (-12.6\%)\end{tabular}  & \begin{tabular}[c]{@{}c@{}} .759\\ (-11.3\%)\end{tabular}   & -                                                          & -                                                         &    -                                                       & -                                                          \\ \bottomrule
\end{tabular}%
}
\caption{F1 score of two representative entity typing models on OntoNotes testing samples with(out) distinct biases and their perturbations.}
\label{tab:diagnosing_ontonotes}
 \end{minipage}
 \begin{minipage}{\textwidth}

\setlength{\tabcolsep}{4pt}
\resizebox{\textwidth}{!}{%
\begin{tabular}{@{}lcccccc|cccccc@{}}\toprule
\multirow{2}{*}{Approach} & \multicolumn{6}{c}{BiLSTM}                                                                                                                                & \multicolumn{6}{c}{MLMET}                                                                                                      \\\cmidrule(r){2-13}
                          & U-Prec.             & U-Rec.                & U-F1                    & A-Prec.             & A-Rec.                & A-F1                    & U-Prec.    & U-Rec.                & U-F1           & A-Prec.    & A-Rec.                & A-F1                    \\\midrule
No Debiasing              &    \textbf{.803}              &      .744          &  \textbf{.773}                 &       .708             &     .609              &  .655                & \textbf{.890} &   .822               &    \textbf{.855}     & .805 &  .706                &   .753               \\\midrule
Augmentation (ours)       & .782           &  \textbf{\textit{.752}}    &   .767      &\textbf{\textit{.781}}  & \textbf{\textit{.711}} &      \textbf{\textit{.745}}    &.777 & \textbf{\textit{.832}}       & .803 & \textbf{\textit{.828}}      &\textbf{\textit{.846}}  & \textbf{\textit{.837}}\\\bottomrule
\end{tabular}%
}
\caption{Effectiveness of the proposed counterfactual augmented approach on two representative entity typing models when testing on OntoNotes test set (U-) and our counterfactual augmented test set (A-).}
\label{tab:mitigation_ontonotes}

 \end{minipage}
\end{table*}

\begin{table}[t!]
\setlength{\tabcolsep}{3pt}
\resizebox{\linewidth}{!}{%
\begin{tabular}{@{}lcccccc@{}}\toprule
Approach& U-Prec.             & U-Rec.                & U-F1                    & A-Prec.             & A-Rec.                & A-F1 \\\midrule
 \multicolumn{7}{c}{\textbf{LabelGCN}} \\\cmidrule(l){2-4} 
No Debiasing              & .498                &   .283              &   .361             &      .503              &  .247                & .332\\\midrule
AFLite                    &\textbf{\textit{.536}}  &      .238           &    .329               &  \textbf{\textit{.529}}     &     .202             &.292 \\
POE                       &   .407                &  \textit{.327}        & \textit{.363} &   .379       &   \textit{.301}        & \textit{.335} \\
Focal                     & .185                  &\textbf{\textit{.439}}  &  .260         & .193                &  \textbf{\textit{.392}}         & .259\\
Learned-mixin             &  .420                  &    \textit{.316}     &  .361        &    .353     &  \textit{.311}      &.331 \\
Learned-mixin+H           &  .225                &    \textit{.398}       &  .287                 &    .212                &  \textit{.339}         & .261\\
Contrastive (CE)          &  .483                  &    \textit{.286}      &   .359       &  .479       &  \textit{.272}       &\textit{.347} \\
Contrastive (Cosine)      &   .453                 &   \textit{.285}      &  .350        &\textit{.516}  &   \textit{.269}        &\textit{.353} \\
Counterfact. Inf.            & .467                   &   \textit{.309}       &  \textbf{\textit{.372}}        &     .403    & \textit{.291}                   &\textit{.338} \\\midrule\midrule
Augmentation (ours)       &  .484        &        \textit{.289} & \textit{.362}         &    \textit{.524}       & \textit{.274} &\textbf{\textit{.360}}\\\midrule
 \multicolumn{7}{c}{\textbf{Box4Types}} \\\cmidrule(l){2-4} 
No Debiasing              & .528                &       .388          &  .448              &          .469          &    .358              & .406\\\midrule
AFLite                    & \textbf{\textit{.531}} &\textit{.400}                 &  \textit{.456}                 & \textit{.473}       &     \textit{.360}             &\textit{.409} \\
POE                       &  .410                 &   \textbf{\textit{.468}}       & .437 &       .347    &   \textit{.433}       & .385 \\
Focal                     &  .407                 &\textit{.467}  &  .435         & .347                &  \textit{.435}         &.386 \\
Learned-mixin             &    .508                &   \textit{.415}      & \textit{.457}         &    .448     &  \textit{.393}      &\textit{.419} \\
Learned-mixin+H           & .463                 &     \textit{.440}      & \textit{.451}                  & .403                   &     \textit{.406}      &.404 \\
Contrastive (CE)          &    .443                &     \textit{.459}    &      \textit{.451}    & .371        &   \textit{.563}      & \textit{.447}\\
Contrastive (Cosine)      &  .472                  &   \textit{.444}      &    \textit{.458}      &.437  &  \textbf{\textit{.499}}         & \textit{.466}\\
Counterfact. Inf.            &  \textit{.529}                  & \textit{.394}         &     \textit{.452}     &     .422    & \textit{.382}                   &.401 \\\midrule\midrule
Augmentation (ours)       &  .521        &     \textit{.410}    & \textbf{\textit{.459}}         &    \textbf{\textit{.504}}       & \textit{.484} &\textbf{\textit{.494}}\\\midrule
 \multicolumn{7}{c}{\textbf{LRN}} \\\cmidrule(l){2-4} 
No Debiasing              &\textbf{.611}                 &           \textbf{.334}      &  \textbf{.432}              &       \textbf{.703}             &   .343               & .461\\
Augmentation (ours)       &    .553      &    .328     & .412         &  .619         &\textbf{\textit{.389}}  &\textbf{\textit{.478}}\\\bottomrule
\end{tabular}%
}
\caption{Effectiveness of different debiasing approaches on remaining entity typing models when testing on UFET test set (U-) and our counterfactual augmented test set (A-). Note that \emph{LRN} predicts types in an autoregressive generative way, it does not provide a fixed logit for each label, hence we can not apply logit-based debiasing approaches to help \emph{LRN} mitigate spurious correlations.}
\label{tab:mitigation_cont}
\end{table}

\end{document}